\documentclass[journal]{IEEEtran}
\AtBeginDocument{%
  }
\usepackage{amsmath}
\usepackage{url}
\usepackage{flushend}
\usepackage{hyperref}
\hypersetup{
    colorlinks=true,
    linkcolor=blue,
    filecolor=blue,      
    urlcolor=blue,
    citecolor=cyan,
}
\usepackage{cite}
\usepackage{pifont}
\usepackage{booktabs}
\usepackage{colortbl}
\usepackage{color}
\usepackage[table]{xcolor}
\newcommand{\gray}{\cellcolor{gray!20}}
\usepackage{hhline}
\usepackage{comment}
\usepackage{multirow}
\usepackage{graphicx}
\usepackage{subfigure}
\usepackage{textcomp}
\definecolor{mygreen}{RGB}{30,119,131}
\definecolor{myred}{RGB}{155,0,18}
\definecolor{myyellow}{RGB}{255,205,49}
\usepackage{bm}

\usepackage{url}
\usepackage{enumitem} 
\usepackage{rotating}
\usepackage{makecell}
\usepackage{hhline}
\begin{document}
\title{FakeScope: Large Multimodal Expert Model for Transparent AI-{G}enerated Image Forensics}
\author{Yixuan~Li,~\IEEEmembership{Member,~IEEE}, Yu~Tian, Yipo~Huang, Wei~Lu,~\IEEEmembership{Member,~IEEE}, Shiqi~Wang\IEEEauthorrefmark{2},~\IEEEmembership{Senior Member,~IEEE}, Weisi Lin,~\IEEEmembership{Fellow,~IEEE} and Anderson Rocha,~\IEEEmembership{Fellow,~IEEE} 
\thanks{Y. Li, Y. Tian and S. Wang are with the College of Computing, City University of Hong Kong, Hong Kong SAR (e-mail: yixuanli423@gmail.com; ytian73-c@my.cityu.edu.hk; shiqwang@cityu.edu.hk). 

Y. Huang is with School of Data Science and Artificial Intelligence, Chang’an University, Xi’an 710064, China (e-mail: huangyipo@hotmail.com)

W. Lu is with the School of Computer Science and Engineering, Ministry of Education Key Laboratory of Information Technology, Guangdong Province Key Laboratory of Information Security Technology, Sun Yat-Sen University, Guangzhou 510006, China (e-mail: luwei3@mail.sysu.edu.cn).

W. Lin is with the College of Computing and Data Science, Nanyang Technological University, Singapore (e-mail: wslin@ntu.edu.sg) 

A. Rocha is with the Artificial Intelligence Lab. (Recod.ai) at the University of Campinas, Campinas 13084-851, Brazil (e-mail: arrocha@unicamp.br)

\IEEEauthorrefmark{2}Corresponding author: Shiqi Wang.}
\thanks{The research involving human subjects was approved by the Research Committee of the City University of Hong Kong (HU-STA-00001048).}
}
\markboth{Under review}%
{Shell \MakeLowercase{\textit{et al.}}: Bare Demo of IEEEtran.cls for IEEE Computer Society Journals}
\maketitle
\begin{abstract}
The rapid and unrestrained advancement of generative artificial intelligence (AI) presents a double-edged sword. While enabling unprecedented creativity, it also facilitates the generation of highly convincing content, undermining societal trust. As image generation techniques become increasingly sophisticated, detecting synthetic images is no longer just a binary task—it necessitates explainable methodologies to enhance trustworthiness and transparency. However, existing detection models primarily focus on classification, offering limited explanatory insights. To address these limitations, we propose \textbf{FakeScope}, an expert large multimodal model (LMM) tailored for AI-generated image forensics, which \textbf{not only identifies synthetic images with high accuracy but also delivers rich query-contingent forensic insights}. At the foundation of our approach is \textbf{FakeChain}, a large-scale dataset containing structured forensic reasoning based on visual trace evidence, constructed via a human-machine collaborative framework. Then we develop \textbf{FakeInstruct}, the largest multimodal instruction tuning dataset to date, comprising two million visual instructions that instill nuanced forensic awareness into LMMs. Empowered by FakeInstruct, FakeScope achieves state-of-the-art performance in both closed-ended and open-ended forensic scenarios. It can accurately distinguish synthetic images, provide coherent explanations, discuss fine-grained forgery artifacts, and suggest actionable enhancement strategies. Notably, despite being trained exclusively on \textbf{qualitative} hard labels, FakeScope demonstrates remarkable zero-shot \textbf{quantitative} detection capability via our proposed token-based probability estimation strategy. Furthermore, it shows robust generalization across unseen image generators and performs reliably under in-the-wild scenarios. The data, model, and demo will be publicly released on \url{https://github.com/Yixuan423/FakeScope}.
\end{abstract}




\section{Introduction}
\IEEEPARstart{W}ith the rapid development of image generation techniques, AI-synthesized images are becoming increasingly indistinguishable from genuine ones~\cite{yan2024sanity,zou2025survey}. While this prosperity has driven innovation, it also introduces significant cybersecurity risks, including misinformation, propaganda, and fraud. In response, substantial efforts have been made to detect AI-generated content in recent years~\cite{ojha2023towards,cozzolino2024raising,cozzolino2024zero,ricker2024aeroblade}. However, as generative models continue to evolve, concerns over the reliability and trustworthiness of detection systems in real-world applications persist. Addressing these challenges requires enhancing the transparency of AI-generated image detection models. We advocate for the potential in which image synthesis detection goes beyond binary classification, incorporating explainability to provide contextual insights into authenticity, assisting trustworthiness.
\begin{figure}[!tbp]
\centering
\includegraphics[scale=0.5]{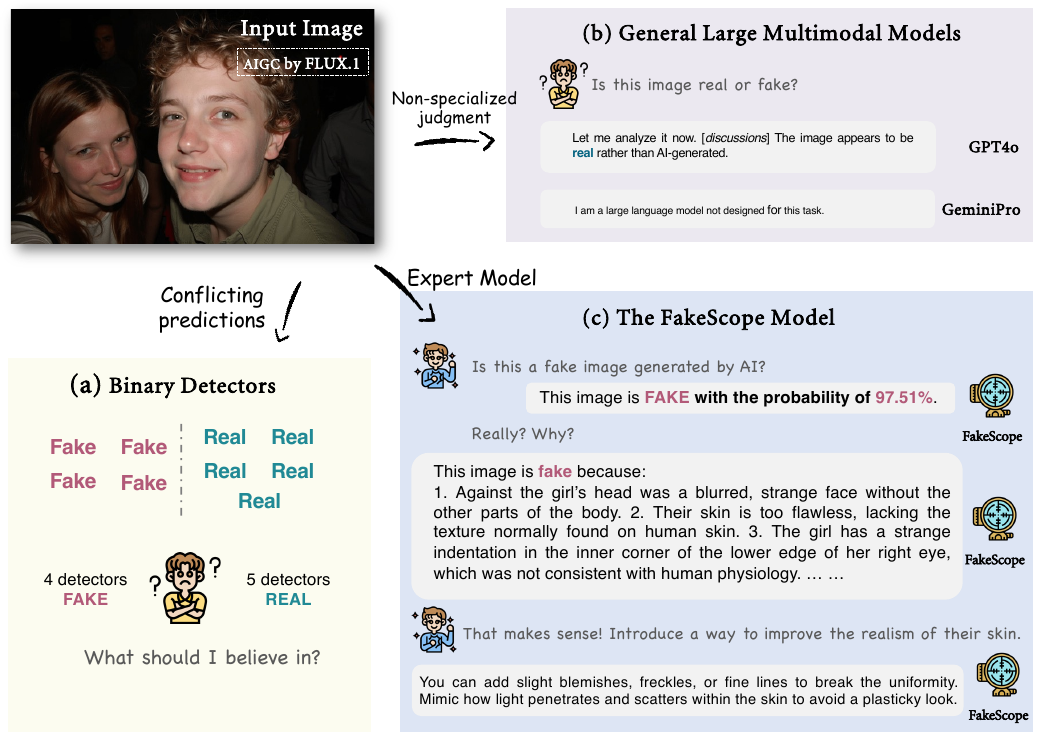}
\caption{\textbf{Comparative illustration of image forensic model evolution.} (a) Binary detectors; (b) general LMMs; and (c) FakeScope, which translates visual abnormalities into human-understandable cues, much like trace evidence in forensics, using natural language.}
\label{fig:intro}
\end{figure}
\begin{figure*}[!tbp]
\centering
\includegraphics[scale=0.83]{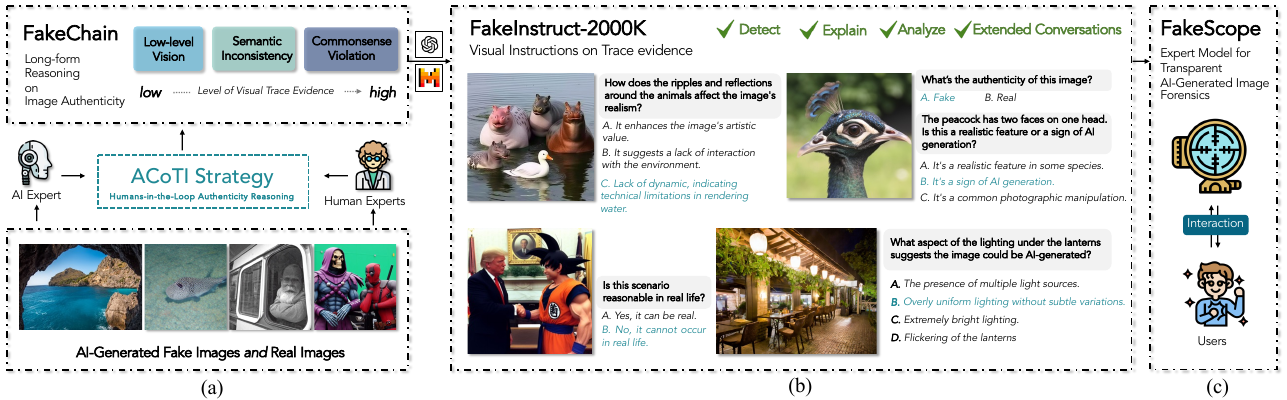}
\caption{Contributions of this work. (a) FakeChain dataset, containing reasoning on image authenticity, constructed via the proposed \textit{ACoTI} strategy (Sec.~\ref{fakechain}); (b) FakeInstruct dataset, containing two million visual instructions concerning image forensic knowledge (Sec.~\ref{sec:fakeinstruction}); (c) FakeScope model, the expert model for transparent AI-generated image forensics, capable of multi-dimensional forensic capabilities (Sec.~\ref{sec:fakescope}).}
\label{fig:overview}
\end{figure*}

Until recently, AI-generated image detection was predominantly framed as a binary classification task. Conventional approaches rely on unimodal architectures such as convolutional neural networks (CNNs) and vision transformers (ViTs) to classify images as either real or synthetic, significantly aiding in mitigating the spread of synthetic content~\cite{wang2020cnn,durall2020watch,frank2020leveraging}. However, these early methods largely overlooked the need for transparency, offering little more than abstract heatmaps derived from deep neural network features as visual explanations. Recent research has shown that AI-generated images increasingly pass sanity checks by humans and machines~\cite{li2024fakebench,yan2024sanity}, leading to human-machine or machine-machine disagreements. As illustrated in Fig.~\ref{fig:intro}(a), different detectors can yield inconsistent decisions, significantly reducing practical utility. In contrast, in Fig.\ref{fig:intro}(c), a transparent\footnote{Following~\cite{gunning2019xai,liu2022trustworthy}, \textit{transparency} herein refers to a composite spectrum including explainability, traceability, and justifiability.} forensic model can go beyond binary decisions by offering visual-evidence-based justifications, supporting query-driven forensic analysis, and estimating authenticity probabilities in a unified and interpretable manner. Such capabilities are key to enabling human oversight, mitigate detection biases, and foster better confidence in AI-generated content forensics~\cite{gunning2019xai,zou2025survey}.

With the advent of multimodal models, vision-language approaches have started to emerge~\cite{sun2023towards,huang2024ffaa,zhang2024common,chen2025mathcalxdfd,guo2025rethinking,wen2025spot,zhou2025aigi}. Zhang \textit{et al.}~\cite{zhang2024common} introduced textual template descriptions to improve explainable deepfake face detection. Sun \textit{et al.}~\cite{sun2023towards} formulated brief descriptions by integrating forgery regions and candidate types using a predefined template. Li \textit{et al.}~\cite{li2024fakebench} and Ye \textit{et al.}~\cite{ye2024loki} explored the capabilities of large multimodal models (LMMs) in the forensic analysis of AI-generated images. In \cite{xu2024fakeshield,liu2024forgerygpt,huang2024sida}, LMMs are incorporated to localize tampering forgeries and exhibit substantial effectiveness. Besides, LMMs are employed to assist in improving detection accuracy. In \cite{huang2024ffaa} and \cite{chen2025mathcalxdfd}, conventional detectors are combined with LMMs to improve detection accuracy. Wen \textit{et al.}~\cite{wen2025spot} train a binary classifier with ViT intermediate features and use LMMs to explain the detection result. These studies have established a concrete foundation for transitioning from unimodal to multimodal approaches in AI-generated image forensics. Nevertheless, due to the inherent limitations of LMMs as classifiers~\cite{zhang2024visually}, previous approaches have struggled to achieve high detection performance and transparency simultaneously. Most existing works rely on separate strategies and training procedures, such as introducing additional classifier heads or independently trained explanation modules, to fulfill different forensic demands. Such fragmented pipeline and weak coupling can lead to disconnected probabilistic reasoning space across tasks and hindering the model from reusing shared forensic knowledge, further constraining the model’s potential. This motivates the need for a unified forensic expert model for AI-generated images capable of performing multiple forensic tasks in diversified scenarios.

Inspired by the insights, we present \textbf{FakeScope}, a unified expert model for transparent and versatile AI-generated image forensics. It leverages the remarkable cross-modality ability of generally pretrained LMMs while being systematically enhanced for forensic awareness. Specifically, it is designed with three key objectives: (1) comprehension of visual trace evidence and authenticity; (2) faithful response to diverse user queries; and (3) qualitative and quantitative detection without model extension. To achieve these, we collect and leverage domain-specific knowledge to improve LMMs' forensic awareness of AI-generated images. Our emphasis is therefore on output-level image forensic transparency rather than on modeling the causal interpretability of internal LMM mechanisms. As shown in Fig.~\ref{fig:overview}(a), we first construct \textbf{FakeChain}, a large-scale multimodal dataset containing long and comprehensive reasoning on image authenticity as a forensic knowledge source, developed via a cost-efficient human-in-the-loop strategy. We further enrich FakeChain to create \textbf{FakeInstruct}, a dataset comprising \textbf{two million} visual instructions to enhance the diverse forensic abilities of LMMs (Fig.~\ref{fig:overview}(b)). Extensive experiments demonstrate that FakeInstruct consistently improves the forensic capabilities of general LMMs, and the proposed \textbf{FakeScope} (Fig.~\ref{fig:overview}(c)) exhibits state-of-the-art (SOTA) performance in detecting, explaining, analyzing, and discussing AI-generated images. In particular, it generalizes well to unseen image content, generation models, and user-created content. Our core contributions are as follows.
\begin{itemize}
    \item We present \textbf{FakeChain}, a large-scale multimodal dataset with structured reasoning from visual trace evidence to image authenticity. It is built with our proposed \textit{\underline{A}nthropomorphic \underline{C}hain-\underline{o}f-\underline{T}hought \underline{I}nference} (\textit{ACoTI}) scheme. This cost-effective human-machine collaboration strategy requires minimal human annotation while ensuring data reliability.
    \item We develop \textbf{FakeInstruct}, the first million-scale visual instruction dataset for enhancing LMMs' forensic capabilities in AI-generated image detection, exhibiting consistent enhancement to diverse base models.
    \item We propose \textbf{FakeScope},  a multimodal forensic expert built on FakeInstruct for transparent AI-generated image detection. It unifies detection and query-based forensic analysis, and further incorporates a zero-shot token-based strategy for probability estimation. Extensive experiments validate its superiority in detection and transparency-related tasks, outperforming other LMMs and binary models across diverse datasets.
\end{itemize}
\section{Related Work}
In this section, we first outline the progression of AI-generated image detection techniques, followed by an overview of recent developments in LMMs.

\subsection{AI-generated Image Detection}

The detection of AI-generated images aims to differentiate fake images (synthesized by generative AI algorithms) from real ones (created or captured by humans). This problem is commonly formulated as a binary classification task. Preexisting approaches generally incorporate forensic patterns to classify, which can be categorized into the flaw-based~\cite{zhong2023rich,ciftci2020fakecatcher} and fingerprints-based~\cite{wang2020cnn,durall2020watch,frank2020leveraging,wolter2022wavelet} according to their forensic ground. The flaw-based methods capture the intricate details in low (\textit{e.g.}, texture~\cite{zhong2023rich}) or high (\textit{e.g.}, physical rules~\cite{ciftci2020fakecatcher,sarkar2024shadows}) semantic levels. At the same time, the fingerprints-based detectors identify the invisible leftover patterns of generators (\textit{e.g.}, frequency artifacts~\cite{wang2020cnn,durall2020watch,frank2020leveraging,wolter2022wavelet}). Moreover, investigations on model's generalization ability towards unseen ~\cite{liu2022detecting,jeong2022fingerprintnet,cozzolino2024raising,ojha2023towards,cozzolino2024zero,ricker2024aeroblade} and generator attribution~\cite{sha2023fake,wang2024did} have emerged. Some of them are assumed to be only accessible to real images and utilize generators (like an autoencoder) to simulate generative artifacts~\cite{zhang2019detecting,frank2020leveraging,jeong2022fingerprintnet,liu2022detecting}. Although remarkable, current fake image detection models still fall short in interpretability towards humans~\cite {lin2024detecting}, which makes the detection results hard to explain for the general populace and weakens the credibility of the model output. Therefore, developing an expert dedicated to fake image detection with high detection accuracy and transparency is urgently needed.

\subsection{Large Multimodal Models}
Typical LMMs comprise a modality encoder, a large language model (LLM), with a modality interface for cross-modal connection \cite{dong2024internlm, huang2024aesBench}. The specialty of LMMs lies in their ability to effectively receive, process, and generate multimodal information. Currently, LMMs are developing by leaps and bounds and have exhibited remarkable visual-language capabilities on par with or even surpassing the human, including the closed-source models (\textit{e.g.}, GPT-4~\cite{gpt4}, Gemini~\cite{team2023gemini}, and Claude~\cite{claude3}) and the open-source ones (\textit{e.g.}, LLaVA~\cite{liu2024visual}, mPLUG-Owl~\cite{ye2023mplug}, Deepseek~\cite{lu2024deepseek}). Furthermore, to adapt general-purpose LMMs to specialized fields, recent researchers proposed various expert models based on visual instruction tuning ~\cite{liu2023mmbench,lu2022learn,du2023makes,liu2024visual}. Despite great progress, recent studies~\cite{li2024fakebench} indicate that LMMs' forensic capabilities are still unsatisfactory, especially when identifying fine-grained forgery aspects. 
In this work, to infuse the forensic abilities of AI-generated images into foundational LMMs, we construct the first visual instruction tuning dataset centering on fake image detection, the \textbf{FakeInstruct}, aiming at a unified multimodal paradigm based on LMMs for a more transparent image forensics. 

\section{FakeChain: Reasoning on Image Authenticity}
\label{fakechain}
In this section, we elaborate on the construction of \textbf{FakeChain}, a large-scale dataset of visual trace-based authenticity reasoning. FakeChain serves as the primary source of forensic knowledge and further supports the construction of the instruction dataset and the expert model.

\begin{figure*}[!tbp]
\centering
\includegraphics[scale=0.88]{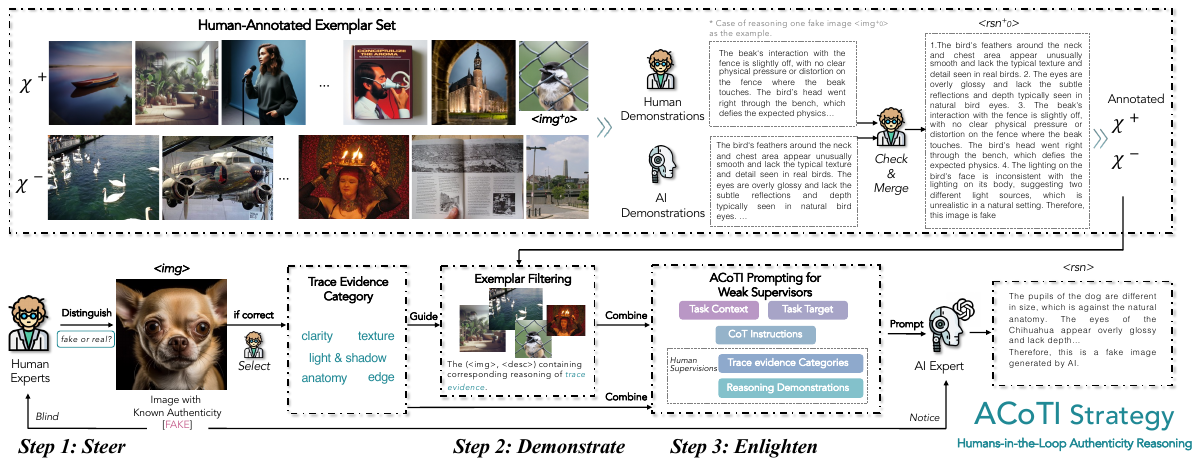}
\caption{Pipeline of the proposed \textit{ACoTI} strategy for obtaining the reasoning on trace evidence under the human-machine collaboration scheme. \textit{ACoTI} is cost-efficient that adopts the human-in-the-loop methodology to incorporate little human labor to \textit{demonstrate, steer,} and \textit{enlighten} non-perfect AI experts to generate qualified reasoning data. This semi-automatic strategy unburdens human labor, leverages LMM's abundant inner knowledge, and ensures reliability simultaneously.}
\label{fig:acoti}
\end{figure*}

We first prepare images (Sec.~\ref{sec:imagecollection}) and then collect authenticity reasoning. While human annotation remains the most intuitive approach for training LMMs~\cite{wu2023q,liu2024visual}, collecting fine-grained authenticity reasoning at scale is highly labor-intensive. To reduce this cost, we adopt a human-machine collaborative strategy (Fig.~\ref{fig:acoti}) that uses limited human annotations as weak supervision in three steps:

\noindent 1. \textit{Steer}, which guides reasoning with human-annotated trace-evidence categories;

\noindent 2. \textit{Demonstrate}, which provides human reasoning exemplars;

\noindent 3. \textit{Enlighten} which elicits visual reasoning from strong models through few-shot chain-of-thought prompting.

Building on this scheme, we first collect human expert feedback (Sec.~\ref{sec:humanannotation}) and propose \textit{\underline{A}nthropomorphic \underline{C}hain-\underline{o}f-\underline{T}hought \underline{I}nference} (\textit{ACoTI}) strategy (Sec.~\ref{sec:acoti}) to \textit{enlighten} strong models under the \textit{steer} of a few human annotations assisted by human reasoning \textit{demonstrations}. With \textit{ACoTI}, we construct the FakeChain dataset. Details are elaborated below.

\vspace{-1.1em}
\subsection{Image Preparation}
\label{sec:imagecollection}
\textbf{FakeChain} contains a equal number of AI-generated images and real images, totaling 47,594. Unlike the one-to-all generalization setting in \cite{liu2022detecting,jeong2022fingerprintnet,cozzolino2024raising,ojha2023towards,cozzolino2024zero,ricker2024aeroblade}, FakeChain is deliberately designed for high diversity to cover a wide range of generative patterns. In particular, AI-generated images function as the knowledge source of forensic trace evidence and are sampled from multiple sources. The generators span 17 types, including GAN-based, diffusion-based, auto-regressive, and proprietary models. Specifically, we choose representative ProGAN~\cite{karras2017progressive}, StyleGAN~\cite{karras2019style}, BigGAN~\cite{brock2018large}, CogView2~\cite{ding2022cogview2}, ADM~\cite{dhariwal2021diffusion}, IF~\cite{deepfloydif}, FuseDream~\cite{liu2021fusedream}, VQDM~\cite{gu2022vector}, Glide~\cite{nichol2021glide}, SD~\cite{rombach2022high}, SDXL~\cite{PodellELBDMPR24}, FLUX.1~\cite{flux}, Firefly~\cite{adboefirefly}, Dalle2~\cite{ramesh2022hierarchical}, Dalle3~\cite{dalle3}, Wukong~\cite{wukong} and Midjourney~\cite{midjourney}. Also, we collect 23,797 real images from DIV2K~\cite{agustsson2017ntire}, RAISE~\cite{dang2015raise}, and ImageNet~\cite{deng2009imagenet}, ensuring content diversity. The details are provided in the supplementary material (SM). 
\subsection{Human Expert Feedback}
\label{sec:humanannotation}
As shown in Fig.~\ref{fig:acoti}, we require a few human annotations in \textit{step1-steer} and \textit{step2-demonstrate}, including the trace evidence category and reasoning demonstration. Therefore, we conduct a controlled in-lab experiment involving 30 trained human experts experienced in image generation, ensuring qualified understanding of generative patterns. In \textit{step1-steer}, experts first distinguish whether an image is AI-generated and then select supporting categories. This ensures only correct annotations contribute to the dataset. In practice, we provide a category set for reference $\mathcal{F}$ =\ \{\textit{texture, edge, clarity, light\&shadow, anatomy, layout, symmetry, reflection, perspective, physics, shape, theme, content deficiency, distortion, unrealistic, overall hue}\}~\cite{li2024fakebench}, covering multiple levels of forensic evidence. To avoid overly broad selections, annotators are instructed to choose only verifiable cues rather than all plausible categories. Practically, each image receives at most six categories. Every image is annotated by two experts, and only the intersection of their selections is kept to reduce annotator-specific bias. Additionally, experts can provide authenticity-related observations at will, such as specific affected regions or relevant commonsense cues. In \textit{step 2-demonstrate}, we fix 50 fake and 50 genuine images and employ experts to annotate manually with detailed reasoning. These annotations constitute the exemplar set $\{{\chi}^{+}, {\chi}^{-}\}$ shown in upper Fig.~\ref{fig:acoti}, where $^{+}$ and $^{-}$ denote AI-generated and genuine images respectively. To ensure coverage, we manually guarantee that each category in $\mathcal{F}$  emerges in at least three exemplars. This process establishes the human supervision needed for subsequent reasoning collection.
\begin{figure*}[!tbp]
\centering
\includegraphics[scale=0.88]{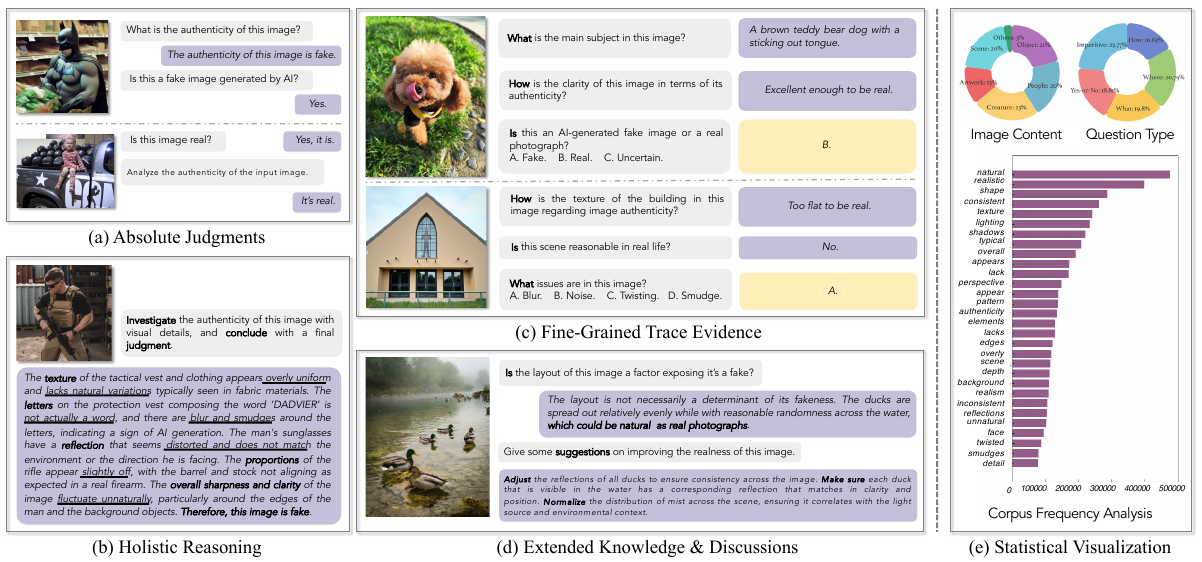}
\caption{The composition of the \textbf{FakeInstruct}, which is derived from the FakeChain dataset, containing 47K visual instructions on absolute authenticity judgments, 95K instructions on holistic reasoning, 715K on fine-grained visual trace evidence, and 1190K on extended knowledge and discussions. The million-scale diversified visual instructions of FakeInstruct enable LMMs with a broad and fine-grained understanding of image authenticity, ensuring LMMs are equipped to handle diverse forensic tasks with both interpretability and accuracy.}
\label{fig:fakeinstruct}
\end{figure*}
\subsection{Reasoning on Visual Trace Evidence}
\label{sec:fakechain}
\subsubsection{Task Definition}
\textbf{FakeChain} encodes forensic knowledge as a \textit{materialized reasoning process} that links exhaustive description of the appearance and location of trace evidence to an authenticity conclusion. Compared with free-form descriptions, this structured format has two advantages: \textbf{(1)} it allows for a precise and detailed reflection of the aspects humans focus on when assessing authenticity; and \textbf{(2)} it follows the pattern of human decision-making. 
By organizing supervision in this way, FakeChain can help to improve human-centered transparency. We emphasize, however, that this materialized reasoning should not be equated with the actual causal basis underlying the model’s authenticity judgment, but rather as a post-hoc explanation.
\subsubsection{ACoTI Strategy}
\label{sec:acoti}
To obtain large-scale reasoning data cost-effectively, we propose a semi-automatic scheme for extracting forensic knowledge from AI experts with marginal human supervision: \textit{\underline{A}nthropomorphic \underline{C}hain-\underline{o}f-\underline{T}hought \underline{I}nference} (\textit{ACoTI}) strategy. As illustrated in Fig.~\ref{fig:acoti}, \textit{ACoTI} follows a human-machine collaborative framework in which humans participate only in the first two stages by providing partial trace-evidence information. This method follows a \textit{Steer}-\textit{Demonstrate}-\textit{Enlighten} pipeline, detailed as follows. 

\textit{\textbf{Step 1. Steer}}: For the $i$-th candidate image, we use the human-annotated trace-evidence category set $\mathcal{F}_i$ to steer the reasoning process. In practice, $\mathcal{F}_{i}$ is carried into the next two stages so that reasoning remains grounded in forensic evidence rather than drifting to irrelevant content.

\textit{\textbf{Step 2. Demonstrate}}: To further reduce hallucination, we provide human-written demonstrations as in-context examples, ~\cite{lampinen2022can}. Considering the trade-off between context length and efficiency, we adaptively select a minimum covering set~\cite{balas1972set} ${\chi}_{i}$ from the exemplar set $\{{\chi}^{+}, {\chi}^{-}\}$ according to $\mathcal{F}{i}$ of the target image. Formally, we solve the following optimization:
\begin{equation}
    \begin{aligned}
        &\chi_{i} = \arg \min_{\chi' \subseteq \{\chi^+ \cup \chi^-\}} |\chi'|, \\
        &\text{\textit{s.t.}} \quad L(\chi') = \mathcal{F}_i, \ \chi' \cap \chi^+ \neq \emptyset, \ \chi' \cap \chi^- \neq \emptyset,
    \end{aligned}
\end{equation}
where $L(\cdot)$ maps a subset of exemplars to the union of trace categories they cover. In practice, we select 3 to 5 demonstrations from ${{\chi}^{+}, {\chi}^{-}}$ for each target image to satisfy both coverage and label-diversity constraints. The selection is implemented with a greedy category-covering algorithm with random tie-breaking, yielding a compact yet diverse demonstration set that captures the key forensic attributes while remaining within the context window.

\textit{\textbf{Step 3. Enlighten}}: The core of ACoTI is a structured prompting scheme that guides the teacher model GPT-4V~\cite{gpt4} to produce forensic reasoning. As illustrated in Fig.~\ref{fig:acoti}, we adopt few-shot chain-of-thought prompting\cite{WeiNIPS2022} augmented with human supervision. For each target image, we construct a prompt as a quadruple \textless\textit{context, target, human supervision, CoT prompt}\textgreater\ , where human supervision includes the attribute set $\mathcal{F}_{i}$, selected demonstrations $\chi_{i}$, and ground-truth label $auth_{i}$. A unique design choice is to explicitly provide the authenticity label in the context to anchor the reasoning toward the appropriate answer space~\cite{WeiNIPS2022}. For example, when the attribute set $\mathcal{F}_i$ = \{texture, shadow\} and ground-truth label $auth_i$ = fake, the model is prompted with task context, target, and human supervision containing attribute guidance and filtered demonstrations. The exact prompting structure is provided in the SM.  

In this way, \textit{ACoTI} provides an effective pipeline for distilling forensic knowledge from strong models into structured reasoning data. Compared with manually collecting descriptions from human annotators, \textit{ACoTI} substantially reduces the burden while maintaining data quality. To quantify its effectiveness, we conducted a two-alternative forced choice (2AFC) study on human preference between the outcomes generated with \textit{ACoTI} against vanilla prompting. Six annotators compared 100 randomly sampled images, and \textit{ACoTI} outputs are preferred in 99.17\% of cases, validating it as a cost-efficient strategy for extracting forensic knowledge from imperfect models. Based on this, we apply \textit{ACoTI} to all collected images and organize the resulting data into the predefined reasoning format $rsn$. As illustrated in Fig.~\ref{fig:acoti}, each image is paired with its $rsn$, forming the final dataset. In total, the resulting \textbf{FakeChain} dataset consists of 47,594 tuples of $\{\bm{\mathtt{I}},auth,rsn\}$, where $\bm{\mathtt{I}}$ and $auth$ denote the image and its authenticity label, respectively.

\section{FakeInstruct: Multimodal Forensic Instruction Dataset}
\label{sec:fakeinstruction}
We transform \textbf{FakeChain} into diverse visual instructions to exploit the information embedded in long-form text. As illustrated in Fig.~\ref{fig:overview}(b), we derive instruction-following data with Mistral~\cite{jiang2023mistral} and GPT-4o~\cite{gpt4} in an intermingling manner to enhance linguistic diversity. We take extra precautions to remove textual contamination, prohibiting models from answering without visual information to ensure visual reasoning~\cite{fu2025blink}. As illustrated in Fig.~\ref{fig:fakeinstruct}, the dataset contains \textbf{2M} visual instructions, equipping LMMs to handle a wide range of forensic scenarios. Details are given below.

\textit{\textbf{Part 1. Absolute Authenticity Judgments:}} As shown in Fig.~\ref{fig:fakeinstruct}(a), we design diverse questions on absolute image authenticity to explicitly instill the concepts of \textit{fake} and \textit{real} in LMMs. We use \textit{fake} to denote the AI-generated images, while \textit{real} for the opposite. We prioritize unambiguous authenticity judgments in this subset and thus not include synonyms. In total, this part comprises about 47K instructions.

\textbf{\textit{Part 2. Holistic Reasoning:}} As shown in Fig.~\ref{fig:fakeinstruct}(b), holistic reasoning induces inference on image authenticity to help LMMs connect multi-level visual trace evidence with overall image authenticity. The structured cause-and-effect format also helps activate latent reasoning abilities for distinguishing AI-generated images~\cite{kojima2022large}.This subset includes 95K instructions.

\textit{\textbf{Part 3. Fine-grained Trace Evidence}}: To help LMMs handle diverse forensic queries, we decompose FakeChain into visual instructions in both narrative and multiple-choice (MCQ) formats. This transformation is facilitated by GPT-4o~\cite{gpt4} and Mistral~\cite{jiang2023mistral}, whose complementary strengths improve instructional diversity. As illustrated in Fig.~\ref{fig:fakeinstruct}(c), this subset covers authenticity-related questions in \textit{what}, \textit{how}, \textit{yes-or-no}, and \textit{imperative} forms, targeting perceptual adjectives, questionable visual attributes (\textit{e.g.}, shape), regions of occurrence (\textit{e.g.}, right eye), or contextual knowledge (\textit{e.g.}, shadows in the wrong direction). Each MCQ comprises one correct answer and several plausible misleading distractors, encouraging sensitivity to subtle cues while reducing misconceptions. This part contains about 715K instructions.

\textbf{\textit{Part 4. Extensional Knowledge:}}
\label{fakeinstruct4}
Beyond forensic analysis, we further introduce extensional knowledge through open-ended discussions (Fig.~\ref{fig:fakeinstruct}(d)).  These discussions cover four aspects: origins of perceptual misjudgments arising from trace evidence, suggesting improvements to enhance the realism of generation, inferring potential generator characteristics, and other user-driven inquiries related to authenticity. This component consists of more than 1190K visual instructions.

Finally, about \textbf{2M} visual instructions are included in the FakeInstruct dataset. As shown in Fig.~\ref{fig:fakeinstruct}(e), the image content type and instruction type are quantitatively well-balanced. Qualitatively, Fig.~\ref{fig:fakeinstruct}(c) shows decent diversity and information consistency across instructions.

\section{FakeScope: Large Multimodal Forensic Model}
\label{sec:fakescope}
In this section, we introduce the proposed large multimodal expert model, \textbf{FakeScope}, which leverages state-of-the-art LMMs as its backbone and is further fine-tuned on FakeInstruct to incorporate forensic knowledge. Moreover, we propose a non-trivial probability estimation strategy via token soft-scoring, which could extend the model's \textit{qualitative} authenticity judgments to \textit{quantifiable} probability estimations, without requiring additional explicit training on numerical labels. This probability estimation operates purely at inference as designed, and does not rely on any auxiliary trainable modules. We aim to enable FakeScope to support user queries beyond binary detection and to provide post-hoc, human-facing rationales that improve the accessibility of its decisions. The details are elaborated as follows.
\subsection{Architecture and Supervised Instruction Tuning}
\textbf{FakeScope} follows the LLaVA-style vision-language architecture~\cite{liu2024visual}.
To verify the generality of FakeInstruct, we instantiate FakeScope with two 7B backbones as variants, including LLaVA-v1.5~\cite{liu2024visual} and mPLUG-Owl2~\cite{ye2023mplug}. Given the scale of FakeInstruct, we further conduct supervised visual instruction tuning on this dataset to adapt the pretrained LMMs to image forensic while also preserving the general knowledge obtained from earlier-phase trainings~\cite{liu2024visual}. 
We freeze the vision encoder and fine-tune only the modality projector and language model for efficiency.
\subsection{Probability Estimation via Token Soft-Scoring}
\label{sec:probability}
\begin{figure}[!tbp]
\centering
\includegraphics[scale=1.2]{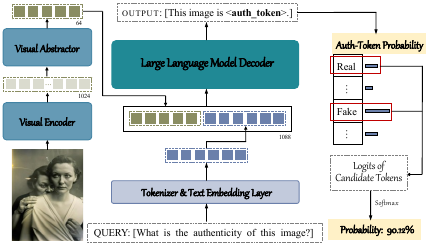}
\caption{Probability estimation in FakeScope. The log probabilities of candidate tokens at the \textless \textit{auth\_token}\textgreater\  position are extracted and normalized using softmax weighting to compute the predicted authenticity probability. The illustration is based on the mPLUG-Owl2~\cite{ye2023mplug} backbone.}
\label{fig:fakescope}
\end{figure}

LMMs typically convey their forensic judgment through selecting the token with the highest logit, which inherently limits their ability to provide numerical probability scores. 
Moreover, since general-purpose LMMs are not explicitly trained as classifiers with numerical labels, direct probability prediction is not a natural formulation. To address this, FakeScope is trained using binary authenticity judgments and incorporates a token \textit{soft-scoring} strategy that operates purely at inference time to convert model outputs into quantitative probability estimates. As depicted in Fig.~\ref{fig:fakescope}, we define \textit{\textbf{fake}} and \textbf{\textit{real}} as anchor tokens and apply softmax normalization over target logits within a standardized prompting framework:
\begin{flushleft}
\texttt{USER:} {\textcolor{myred}{\texttt{<Image> What is the authenticity of this image?}}}

\texttt{ASSISTANT:} {\textcolor{myred}{\texttt{This image is \text{<auth\_token>}.}}}
\end{flushleft}
Probabilities are computed from the logits at the position of \textless \textit{auth\_token}\textgreater, where the candidate token sets are defined as $\mathcal{T} = \{\tilde{t}_j|_{j=0}^{1}\} = \{\tilde{t}_{fake},\ \tilde{t}_{real}\}$, where $\tilde{t}_{(\cdot)}$ represents case variations of the anchor tokens for robustness. The probability $p_i$ of each class is then calculated as follows,
\begin{equation}
\label{softmaxdetect}
    p_j = \frac{e^{\sum_{v \in \tilde{t}_j} z^{(v)}}}{\sum_{j=0}^1 e^{\sum_{v \in \tilde{t}_j} z^{(v)}}},
\end{equation}
where $z^{(v)}$ represents the logit associated with token $v$. Unlike hard token selection, this approach retains nuanced confidence signals that would otherwise be discarded during deterministic text generation. This enables FakeScope to generate \textbf{quantifiable probability estimates}, improving its explainability as a forensic detector.
\begin{table*}[!tbp]
  \centering
  \caption{Publicly available datasets used for evaluation. The {\ddag} notion indicates the uncontaminated content of the dataset, and \# denotes the unit number.}
  \renewcommand{\arraystretch}{1}
\fontsize{7pt}{7pt}\selectfont %
\centering
\setlength{\tabcolsep}{0.8mm}{
    \begin{tabular}{llcclcl}
    \toprule
    \textbf{Purpose}& \textbf{Datasets} & \textbf{Modality} & \textbf{\# Real/ \# Fake} & \textbf{Real Image Source} & \textbf{\# Generators}  &\textbf{Evaluation Criteria}\\
    \midrule
          \makecell[l]{Multimodal Forensic\\Capability} &FakeBench~\cite{li2024fakebench} & Multiple & 3k / 3k & ImageNet, DIV2K, COCO & 10  & \makecell[l]{Detecting, Interpreting, Reasoning,\\Fine-grained Analyzing,\\ Forensic Attributing} \\
    \midrule
          \multirow{2}*{Generalization Capability}&AntiFakePrompt\textsuperscript{\ddag}~\cite{chang2023antifakeprompt} &\multirow{2}*{Single} &42k / 6k       &COCO &11   &\multirow{2}*{Detecting}\\
          &AIGCDetectBenchmark\textsuperscript{\ddag}~\cite{zhong2024patchcraft}&       &74.3k / 74.3k &COCO, LSUN, FFHQ, CelebA       &7\\
    \midrule
          \multirow{2}*{In-the-wild Performance} &WildRF~\cite{cavia2024real} &\multirow{2}*{Single}    & 1.25k / 1.25k      & Reddit, FB, X & Unknown  &\multirow{2}*{Detecting}\\
          &SynthWildX~\cite{cozzolino2024raising} &      & 500 / 1.5k      & X & 3 &\\
    \bottomrule
    \end{tabular}%
    }
  \label{tab:datasets}%
\end{table*}%
\section{Experiments and Analyses}
\subsection{Protocols}
In this section, we present a systematic evaluation of FakeScope, analyzing several key aspects of its performance:
\begin{itemize}
    \item \textbf{Multimodal Forensic Capabilities}: We evaluate to what extent the FakeScope can function as a \textit{transparent} image forensic model (Sec.~\ref{exp1}).
    \item \textbf{In-the-wild Performance}: We assess FakeScope's performance on user-generated images collected from social networks (Sec.~\ref{exp5}).
    \item \textbf{Generalization Capability}: We examine how well FakeScope generalizes to unseen image content and generators (Sec.~\ref{exp3}).
\end{itemize}
Additionally, we perform ablation studies to analyze the impact of different training data components, scales, and inference strategies (Sec.~\ref{exp4}). Detailed experimental setups are provided below.
\subsubsection{Implementation Details}

FakeScope is trained in PyTorch on eight NVIDIA H20-NVLink 98GB GPUs, and evaluated on a single NVIDIA Tesla A100 40GB GPU. All runs use the full-schedule setting with full-parameter fine-tuning of the language model and modality projector, while the vision encoder is frozen for efficiency and stability. Detailed hyperparameters are provided in the SM. Training is performed for one epoch and completes in approximately 34 hours. For each base model, we retain a single checkpoint for all evaluations to ensure fair comparisons. No data augmentation is applied during training.
\subsubsection{Evaluation Datasets}
We evaluate \textbf{FakeScope} on five widely used benchmarks for AI-generated image detection. As summarized in Table~\ref{tab:datasets}, we assess multimodal forensic capabilities on FakeBench~\cite{li2024fakebench}, which covers both binary authenticity prediction and open-ended forensic criteria. We further extend FakeBench to FakeBench\textsuperscript{\texttt{MCQ}} with 3,000 close-ended multiple-choice questions on local visual attributes to evaluate fine-grained forensic attribution; details are provided in the SM. To assess generalization to \textit{unseen} content and generation models, we use AntifakePrompt~\cite{chang2023antifakeprompt}, whose fake images are derived from Microsoft COCO~\cite{lin2014microsoft} exclusive in training data. Additionally, for AIGCDetectBenchmark~\cite{zhong2024patchcraft}, we employ the strictly non-overlapping evaluation split by removing all real images from ImageNet~\cite{deng2009imagenet} and fake images whose underlying real-image provenance traces back to ImageNet. This source-level isolation ensures no content is shared across training and testing. For in-the-wild evaluation, we use WildRF~\cite{cavia2024real} and SynthWildX~\cite{cozzolino2024raising}, which consist of user-generated social-media images. Finally, scenarios such as deepfakes (\textit{e.g.}, FF++~\cite{roessler2019faceforensicspp}) and image super-resolution are treated as out-of-distribution (OOD) cases because they typically modify real images rather than performing end-to-end synthesis; we nevertheless include them in the generalization evaluation.

\subsubsection{Baseline Models} 
FakeScope is compared against a broad set of baselines: six open-source LMMs, including InstructBLIP~\cite{li2023blip}, IDEFICS-Instruct~\cite{laurenccon2023obelics}, InternLM-XComposer2-VL~\cite{dong2024internlm}, LLaVA-v1.5~\cite{liu2024visual}, Qwen-VL~\cite{bai2023qwen}, and mPLUG-Owl2~\cite{ye2023mplug}; three proprietary LMMs, namely GPT-4V~\cite{gpt4}, GeminiPro~\cite{team2023gemini}, and Claude3-Sonnet~\cite{claude3}; three data-driven binary detectors, CNNSpot~\cite{wang2020cnn}, UnivFD~\cite{ojha2023towards}, and FreDect~\cite{frank2020leveraging}; and the vision-language contrastive model CLIP-ViT-Large-14~\cite{radford2021learning}. For detection, LMMs are evaluated in both qualitative (textual judgment) and quantitative (probability estimation) settings, while binary detectors are evaluated only in the quantitative setting. To ensure fairness, the data-driven detectors are retrained on the same images as FakeScope.  All LMMs are evaluated using the same prompting templates as FakeScope. These settings remain consistent across all the experiments, unless explicitly stated otherwise.
\begin{table*}[!t]
\caption{Forensic detection results on the FakeBench-FakeClass~\cite{li2024fakebench} under the \textit{\textbf{qualitative}} setting. The detection Acc. (\%) of each model is reported. The first and second best LMMs are highlighted in \textbf{bold} and \underline{underlined}, respectively. The ``Fake'' column reports the average accuracy across all individual  ``Generation Model'' columns, and both the averaged ``Authenticity'' columns and ``Question Type'' columns reflect the ``Overall'' column.}  
\label{tab:fakeclass-result1}
\renewcommand{\arraystretch}{1}
\fontsize{7pt}{7pt}\selectfont %
\centering
\setlength{\tabcolsep}{0.4mm}{
\begin{tabular}{lccccccccccccccc}
\toprule
\textbf{Subcategories}  & \multicolumn{2}{c}{\textbf{Authenticity}}   & \multicolumn{2}{c}{\textbf{Question Type}}  & \multicolumn{10}{c}{\textbf{Generation Model}}   & {\multirow{2}*{\textbf{Overall}}}\\
\cmidrule(lr){1-1}\cmidrule(lr){2-3}\cmidrule(lr){4-5}\cmidrule(lr){6-15}
Model &\emph{Fake} &\emph{Real} &\emph{What} &\emph{Yes/No} &\emph{proGAN} &\emph{styleGAN} &\emph{CogView} &\emph{FuseDream} &\emph{VQDM} &\emph{Glide} &\emph{SD} &\emph{DALL$\cdot$E2}&\emph{DALL$\cdot$E3}&\emph{MJ} \\
\midrule 
\emph{Random guess}  &42.27 &58.03 &51.80 &48.50 &40.33 &36.00 &51.67 &33.67 &38.00 &44.67 &32.33 &47.67 &37.67&60.67 &50.15    \\
\arrayrulecolor{gray}
\midrule
\textit{Human (Best)} &97.00 &93.00 &/&/ &100.00&100.00&100.00&100.00&100.00&100.00&100.00&100.00&100.00&93.00&87.50 \\
\textit{Human (Worst)} &43.00 &13.00 &/&/&50.00&0.00&46.15&77.78&33.33&16.67&41.67&7.14&14.29&0.00&55.00\\
\textit{Human (Overall)} &76.91 &72.12 &/&/&95.47 &58.82  &87.07 &95.90 &63.96 &82.91 &79.41 &77.66 &70.69 &45.59 &74.51 \\
\midrule
GPT-4V \textit{(Proprietary, teacher)} &59.87 &96.20 &76.33 &79.73 &95.00 &66.00 &66.67 &81.33 &36.67 &49.00 &45.67 &44.33 &61.00 &53.00 &78.03\\
GeminiPro \textit{(Proprietary)} &35.83 &99.27 &61.27 &73.83 &45.33 &15.67 &47.33 &52.67 &19.33 &28.00 &35.33 &35.67 &41.00 &38.00&67.50 \\
Claude3 Sonnet \textit{(Proprietary)} &12.00 &98.23 &55.97 &54.27 &7.00 &0.00 &20.33 &13.67 &0.67 &2.67 &15.33 &9.33 &30.33 &20.67 &55.12 \\
\midrule
InternLM-XC.2-vl \textit{(InternLM2-7B)} &32.17 &92.33&60.40 &64.10 &54.00 &12.67 &43.33 &49.00 &19.00 &20.33 &29.33 &29.67 &34.67 &29.67 &62.25\\
InstructBLIP \textit{(Vicuna-7B)} &67.80 &47.67 &65.23 &50.23 &80.67 &65.33 &66.00 &71.33 &68.00 &70.33 &60.33 &61.00 &70.67 &64.33 &57.73\\
Qwen-VL \textit{(Qwen-7B)} &28.57 &84.27 &51.17 &61.17 &45.67 &5.67 &35.00 &37.67 &8.33 &14.33 &31.33 &21.00 &46.67 &40.00 &56.42 \\
IDEFICS-Instruct \textit{(LLaMA-7B)} &24.97 &31.97 &7.00 &49.93 &19.67 &27.33 &24.33 &28.33 &27.67 &24.00 &23.67 &23.67 &24.00 &27.00 &28.47\\
LLaVA-v1.5 \textit{(Vicuna-7B, baseline)} &38.00 &77.40 &55.77 &59.63 &44.33 &17.33 &45.33 &55.33 &23.00 &28.33 &39.67 &0.00 &0.00 &41.00 &57.70\\
mPLUG-Owl2 \textit{(LaMA2-7B, baseline)} &49.60 &93.97 &71.87 &71.70&56.33 &10.00 &72.00 &75.33 &19.33 &32.00 &56.00 &55.00 &64.00 &56.00 &71.78\\
\midrule
\textbf{FakeScope} \textit{(\textbf{Ours}, mPLUG-Owl2)} &\textbf{99.50} &\textbf{99.87} &\textbf{99.73} &\textbf{99.63} &\textbf{100.00} &\textbf{100.00} &\textbf{99.67} &\textbf{100.00} &\textbf{100.00} &\textbf{100.00} &\underline{98.67} &\textbf{100.00} &\underline{97.67} &\textbf{99.00} &\textbf{99.68}\\
\rowcolor{gray!15}
Improvement &{\textcolor{mygreen}{+49.90}} &{\textcolor{mygreen}{+5.90}} &{\textcolor{mygreen}{+27.86}} &{\textcolor{mygreen}{+27.93}} &{\textcolor{mygreen}{+43.67}} &{\textcolor{mygreen}{+90.00}} &{\textcolor{mygreen}{+27.67}} &{\textcolor{mygreen}{+24.67}} &{\textcolor{mygreen}{+80.67}} &{\textcolor{mygreen}{+68.00}} &{\textcolor{mygreen}{+42.67}} &{\textcolor{mygreen}{+45.00}} &{\textcolor{mygreen}{+33.67}} &{\textcolor{mygreen}{+43.00}} &{\textcolor{mygreen}{+27.90}}\\
\textbf{FakeScope} \textit{(\textbf{Ours}, LLaVA-v1.5)} &\underline{98.67} &\underline{97.93} &\underline{98.67} &\underline{97.93} &\underline{99.67} &\underline{98.67} &\underline{99.00} &\underline{99.67} &\underline{96.67} &\underline{97.00} &\underline{99.33} &\textbf{100.00} &\textbf{98.00} &\underline{98.67} &\underline{98.30} \\
\rowcolor{gray!15}
Improvement &{\textcolor{mygreen}{+60.67}} &{\textcolor{mygreen}{+20.53}} &{\textcolor{mygreen}{+42.90}}&{\textcolor{mygreen}{+38.30}} &{\textcolor{mygreen}{+55.34}}&{\textcolor{mygreen}{+81.34}} &{\textcolor{mygreen}{+53.67}}&{\textcolor{mygreen}{+44.34}} &{\textcolor{mygreen}{+73.67}}&{\textcolor{mygreen}{+68.67}} &{\textcolor{mygreen}{+59.66}}&{\textcolor{mygreen}{+100.00}}&{\textcolor{mygreen}{+98.00}}&{\textcolor{mygreen}{+57.67}} &{\textcolor{mygreen}{+40.60}}\\
\arrayrulecolor{black}
\bottomrule
\end{tabular}}
\end{table*}
\subsubsection{Performance Measures}
FakeScope is designed to handle both forensic detection and open-ended analysis, so we evaluate it with a diverse set of metrics. For forensic detection, we consider both qualitative and quantitative settings. In the qualitative setting, classification accuracy is defined as $Acc.=N_c/N_t$, where $N_c$ and $N_t$ are the numbers of correct and total responses. In the quantitative setting, following~\cite{wang2020cnn,ojha2023towards}, we use threshold-independent calibration metrics, including average precision (AP) and area under the ROC curve (AUC), which are more robust to dataset class imbalance. To assess transparency-related performance, we measure how well model responses align with reference answers both conventional word-level metrics (BLEU~\cite{papineni2002bleu}, ROUGE~\cite{chin2004rouge}, and BERTScore~\cite{reimers2019sentence}) and the LLM-as-a-judge strategy~\cite{gu2024survey} to measure semantic-level alignment, following \cite{lu2022learn,liu2024visual,li2024fakebench,wen2025spot}. These scores are normalized and averaged into an overall transparency score, where higher values indicate better performance. Forensic attribution is evaluated by accuracy on FakeBench\textsuperscript{\texttt{MCQ}}. Details on the evaluation measures are further provided in the SM.

\subsection{Multimodal Forensic Capability}
\label{exp1}
\begin{figure*}[!tbp]
\centering
\includegraphics[width=\textwidth]{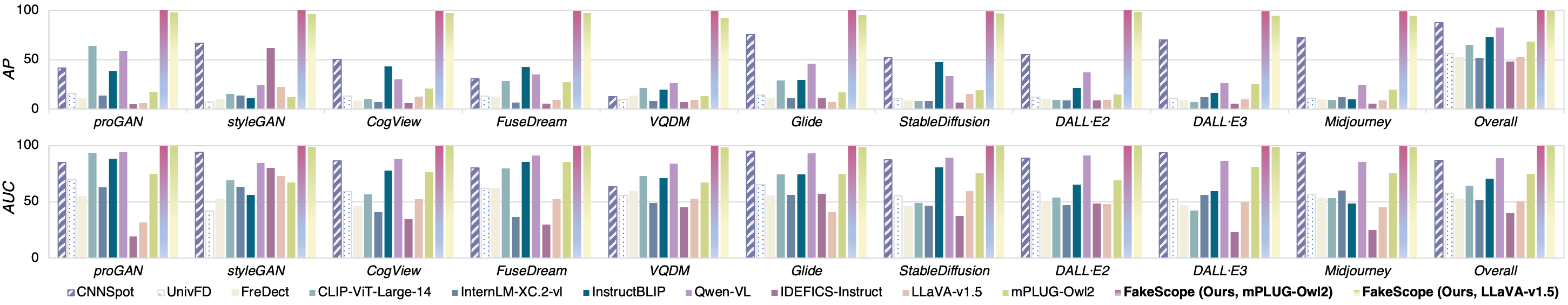}
\caption{Forensic detection results on FakeClass~\cite{li2024fakebench} under the \textit{\textbf{quantitative}} setting. AP and AUC (\%) are reported for each image generation model. The \textbf{FakeScope} far exceeds the other models on each subset and the entire dataset.}
\label{fakeclass-result2}
\end{figure*}
The multimodal forensic capabilities of the proposed \textbf{FakeScope} model are quantitatively assessed on FakeBench containing FakeClass, FakeClue, and FakeQA subsets concerning four ability criteria for LMMs comprehensively defined in \cite{li2024fakebench}, which are stated as follows.
\begin{itemize}
    \item \textbf{\textit{Detecting}}, which evaluates whether forensic classifiers can accurately differentiate AI-generated fake images from real ones;
    \item \textbf{\textit{Interpreting}}, which examines how well LMMs can extract visual evidence to support the \textit{known} authenticity;
    \item \textbf{\textit{Reasoning}}, which assesses whether LMMs can identify generative visual trace evidence and correctly infer the \textit{unknown} image authenticity;
    \item \textbf{\textit{Fine-grained Analyzing}}, which investigates whether LMMs can discuss and analyze specific aspects of telltale visual clues;
    \item \textbf{\textit{Forensic Attributing}}, which investigate whether LMMs can spot specific forensic visual cues in images. 
\end{itemize}
In practice, all LMM responses are generated using the \textit{greedy search} strategy to ensure reproducibility. The quantitative evaluation results for the four criteria are presented below, with qualitative examples provided in SM.
\begin{table*}[!t]
\caption{Multimodal forensic capability comparisons on FakeBench~\cite{li2024fakebench}, including detecting on FakeClass, interpreting on FakeClue-faultfinding, reasoning on FakeClue-inference, and fine-grained forgery analyzing on FakeQA. The -M and -L denote base model mPLUG-Owl2 and LLaVA-1.5, respectively. The B.-1/B.-2 denote BLEU-1/2, R.-L is ROUGE-L, Sim. is BERTScore similarity, and $\alpha$, $\rho$, $\kappa$ correspond to \textit{completeness}, \textit{preciseness}, and \textit{relevance} judged by LLMs. Each column's first- and second-best performances are highlighted in \textbf{bold} and \underline{underlined}.}  
\label{tab:multimodalcapab}
\renewcommand{\arraystretch}{1.1}
\fontsize{6pt}{7pt}\selectfont %
\centering
\setlength{\tabcolsep}{0.2mm}{
\begin{tabular}{lccccccccccccccccccccccccccc}
\toprule
 \textbf{Capabilities} &\multicolumn{8}{c}{\textbf{Interpreting}}&\multicolumn{8}{c}{\textbf{Reasoning}} &\multicolumn{8}{c}{\textbf{Fine-grained Analyzing}}  &\multirow{3}*{\textbf{Overall}}\\
\cmidrule(lr){1-1}
\cmidrule(lr){2-9} \cmidrule(lr){10-17} \cmidrule(lr){18-25}

\multirow{2}*{\textbf{LMMs}} & \multicolumn{4}{c}{\textbf{Word-level}}  & \multicolumn{3}{c}{\textbf{Semantic-level}} &\multirow{2}*{\textbf{Avr.}$\uparrow$} & \multicolumn{4}{c}{\textbf{Word-level}}  & \multicolumn{3}{c}{\textbf{Semantic-level}} &\multirow{2}*{\textbf{Avr.}$\uparrow$}  & \multicolumn{4}{c}{\textbf{Word-level}}  & \multicolumn{3}{c}{\textbf{Semantic-level}} &\multirow{2}*{\textbf{Avr.}$\uparrow$} & \\

\cmidrule(lr){2-5}\cmidrule(lr){6-8} \cmidrule(lr){10-13}\cmidrule(lr){14-16}
\cmidrule(lr){18-21}\cmidrule(lr){22-24}
&\emph{B.-1} &\emph{B.-2} &\emph{R.-L} &\emph{Sim.}  &$\alpha$ &$\rho$ &$\kappa$ & &\emph{B.-1} &\emph{B.-2} &\emph{R.-L} &\emph{Sim.}  &$\alpha$ &$\rho$ &$\kappa$ & &\emph{B.-1} &\emph{B.-2} &\emph{R.-L} &\emph{Sim.}  &$\alpha$ &$\rho$ &$\kappa$ &  &\\
\midrule 
GPT-4V \textit{(Teacher)}  &0.188 &0.092 &0.211 &0.635 &1.872 &1.423 &\underline{1.891}&0.573&0.273 &0.118 &0.225 &0.612&1.392&0.777&1.503&0.460 &0.062 &0.022&0.177&0.455&1.518&1.261&1.869&0.477 &0.503\\
GeminiPro &0.068 &0.029 &0.113 &0.459&1.688&1.024&1.271&0.415&0.168 &0.059 &0.181 &0.485 &0.833&0.561&1.239 &0.331 &0.178 &0.071&0.199&0.365&1.247&0.675&1.686&0.402 &0.383\\
Claude3 Sonnet &0.130&0.042&0.170&0.467&1.516&1.182&1.884&0.483 &0.188&0.062&0.193&0.528&0.928&0.441&1.456 &0.357 &0.038 &0.012 &0.158&0.398&1.254&1.091&1.886&0.429 &0.423\\
\arrayrulecolor{gray}
\midrule
InstructBLIP  &0.207 &0.072&0.206&0.463&1.841 &1.398&1.829&0.541 &0.166&0.065&0.160&0.458 &1.828&0.922&1.546 &0.464 &0.004&0.001&0.007&0.079&0.810&0.321&0.543&0.151 &0.385\\
IDEFICS-Instruct  &0.093&0.032&0.140&0.367&\textbf{1.936}&1.344&1.789&0.501&0.099&0.026&0.148&0.457&1.178&0.450&0.967&0.308 &0.189&0.086&0.231&0.381&1.165&0.456&1.483&0.370 &0.393\\
InternLM-XC.2-vl &0.095&0.032&0.142&0.402&\underline{1.899}&1.260&1.534&0.475&0.108&0.032&0.169&0.492&1.004&0.670&1.167&0.337 &0.146&0.054&0.173&0.366&1.208&0.719&1.623&0.388 &0.400\\
LLaVA-v1.5 &0.143&0.047&0.159&0.466&1.812&0.964&1.621&0.468&0.151&0.051&0.167&0.465&0.905&0.404&1.166 &0.311 &0.128&0.051&0.161&0.360&1.259 &0.757 &1.659&0.394 &0.391\\
Qwen-VL &0.056&0.018&0.096&0.322&1.853&1.149&1.485&0.432&0.130&0.045&0.159&0.469&0.929&0.401&1.104&0.303 &0.076&0.026&0.105&0.241&1.004&0.490&0.802&0.247 &0.327\\
mPLUG-Owl2
&0.099&0.032&0.146&0.455&1.721&0.839&1.415&0.423&0.109&0.034&0.160&0.458&0.843&0.500&1.070&0.296 &0.187&0.091&0.228&0.439&1.089 &0.412 &1.504 &0.368 &0.362\\
\midrule
    \textbf{FakeScope-M} &\textbf{0.434}       &\textbf{0.256}       &\textbf{0.371}       &\underline{0.704}       &1.897       &\underline{1.608}       &\textbf{1.901}       &\underline{0.638}  &\textbf{0.439} &\textbf{0.264} &\textbf{0.383} &\textbf{0.747} &\textbf{1.886} &\textbf{1.265} &\underline{1.775} &\textbf{0.614} &\underline{0.290} &\underline{0.174} &\underline{0.293} &\underline{0.603} &\underline{1.628} &\underline{1.356} &\underline{1.857} &\underline{0.540} &\underline{0.597}\\
\rowcolor{gray!15}
    Improvement &{\textcolor{mygreen}{+0.335}} &{\textcolor{mygreen}{+0.224}} &{\textcolor{mygreen}{+0.225}} &{\textcolor{mygreen}{+0.249}} &{\textcolor{mygreen}{+0.176}} &{\textcolor{mygreen}{+0.769}} &{\textcolor{mygreen}{+0.486}} &{\textcolor{mygreen}{+0.215}} &{\textcolor{mygreen}{+0.330}} &{\textcolor{mygreen}{+0.230}} 
 &{\textcolor{mygreen}{+0.223}} &{\textcolor{mygreen}{+0.289}} &{\textcolor{mygreen}{+1.043}} &{\textcolor{mygreen}{+0.765}} &{\textcolor{mygreen}{+0.705}} &{\textcolor{mygreen}{+0.318}} &{\textcolor{mygreen}{+0.103}} &{\textcolor{mygreen}{+0.083}} &{\textcolor{mygreen}{+0.065}} &{\textcolor{mygreen}{+0.164}} 
 &{\textcolor{mygreen}{+0.539}} &{\textcolor{mygreen}{+0.944}} &{\textcolor{mygreen}{+0.353}} &{\textcolor{mygreen}{+0.172}}&{\textcolor{mygreen}{+0.235}}\\
    \textbf{FakeScope-L} &\underline{0.432}      & \underline{0.254}      &\underline{0.370}       & \textbf{0.708}      & 1.893      & \textbf{1.653}      &1.887   &\textbf{0.640} &\underline{0.437} &\underline{0.262} &\underline{0.381} &\underline{0.745}  &\underline{1.873} &\underline{1.242} &\textbf{1.776} &\underline{0.610} &\textbf{0.304} &\textbf{0.184} &\textbf{0.305} &\textbf{0.611} &\textbf{1.647} &\textbf{1.402} &\textbf{1.879} &\textbf{0.553} &\textbf{0.601}\\
\rowcolor{gray!15}
    Improvement &{\textcolor{mygreen}{+0.289}} &{\textcolor{mygreen}{+0.207}} &{\textcolor{mygreen}{+0.211}} &{\textcolor{mygreen}{+0.242}} &{\textcolor{mygreen}{+0.081}} &{\textcolor{mygreen}{+0.689}} &{\textcolor{mygreen}{+0.266}} &{\textcolor{mygreen}{+0.172}} &{\textcolor{mygreen}{+0.286}} &{\textcolor{mygreen}{+0.211}} &{\textcolor{mygreen}{+0.214}} &{\textcolor{mygreen}{+0.280}} &{\textcolor{mygreen}{+0.968}} &{\textcolor{mygreen}{+0.838}} &{\textcolor{mygreen}{+0.610}} &{\textcolor{mygreen}{+0.299}} &{\textcolor{mygreen}{+0.176}} &{\textcolor{mygreen}{+0.133}} &{\textcolor{mygreen}{+0.144}} &{\textcolor{mygreen}{+0.251}} &{\textcolor{mygreen}{+0.388}} &{\textcolor{mygreen}{+0.645}} &{\textcolor{mygreen}{+0.220}} &{\textcolor{mygreen}{+0.159}} &{\textcolor{mygreen}{+0.210}}\\
\arrayrulecolor{black}
\bottomrule
\end{tabular}
}
\end{table*}
\begin{table}[!tbp]
  \centering
  \renewcommand{\arraystretch}{1.1}
\fontsize{7pt}{7pt}\selectfont %
  \caption{Forensic attribution capability (\%) on FakeBench\textsuperscript{\texttt{MCQ}}. The -M and -L denote base model mPLUG-Owl2 and LLaVA-1.5.}
  \setlength{\tabcolsep}{0.5mm}{
    \begin{tabular}{lcccccccc}
    \toprule
    \multirow{2}[4]{*}{\textbf{Model}}& \multicolumn{4}{c}{\textbf{Question Type}} & \multicolumn{3}{c}{\textbf{Attribution Level}}             & \multirow{2}[4]{*}{\textbf{Overall}} \\
\cmidrule(lr){2-5} \cmidrule(lr){6-8}         & \textit{Yes/No} & \textit{What} & \textit{Where} &\textit{How} & \textit{Low} & \textit{Medium} & \textit{High} \\
    \midrule
    \textit{Random guess} & 50.13      & 37.60      & 24.27      & 26.93      & 34.50                    & 37.80      & 31.90      & 34.73    \\
    \midrule
    GPT-4V (\textit{Teacher}) & \textbf{79.47} & 68.80 & 55.87 & 62.40 & 70.70 & 61.50 &67.60 &66.60\\
    GeminiPro &70.37 &66.20&52.13 &70.80&65.20 &62.60&66.80&64.88\\
    Claude3 Sonnet &68.27 & 67.73 & 66.80 & 65.20 &68.30 &68.80 &63.90& 67.00\\
    \midrule
    InstructBLIP &56.27 &56.00&54.93 &53.60&57.70 &58.40&49.50&55.20\\
    IDEFICS-Instruct &48.67 &49.60 &52.27 &53.47&50.20 &49.70&53.10&51.00\\
    InternLM-XC.2-vl &69.73 &68.13 &66.27&59.47&66.30&65.50 &65.90&65.90\\
    LLaVA-v1.5 &60.40 &62.27 &63.07&61.60 &63.80 &62.60&59.11&61.83\\
    Qwen-VL &64.13&64.93&60.93&60.67 &62.30 &61.80&63.90&62.67\\
    mPLUG-Owl2 &70.53 &66.93 &62.27&64.13 &64.80 &66.30& 66.80 &65.97\\
    \midrule
    FakeScope-M &\underline{75.47} &\textbf{79.73} &\textbf{78.53} &\textbf{80.27}&\underline{76.80} &\textbf{80.50}&\textbf{78.20}&\textbf{78.50}\\
    \rowcolor{gray!15}
    Improvement &{\textcolor{mygreen}{+4.94}} &{\textcolor{mygreen}{+12.80}} &{\textcolor{mygreen}{+16.26}} &{\textcolor{mygreen}{+16.14}} &{\textcolor{mygreen}{+12.00}} &{\textcolor{mygreen}{+14.20}} &{\textcolor{mygreen}{+11.40}} &{\textcolor{mygreen}{+12.53}}\\
    FakeScope-L &\textbf{79.47} &\underline{78.00} &\underline{78.27} &\underline{75.20} &\textbf{82.80} &\underline{73.70}&\underline{76.70} &\underline{77.74}\\
    \rowcolor{gray!15}
    Improvement &{\textcolor{mygreen}{+19.07}} &{\textcolor{mygreen}{+15.73}} &{\textcolor{mygreen}{+15.20}} 
    &{\textcolor{mygreen}{+13.60}} &{\textcolor{mygreen}{+19.00}} &{\textcolor{mygreen}{+11.10}} 
    &{\textcolor{mygreen}{+17.59}} &{\textcolor{mygreen}{+15.91}}\\

    \bottomrule
    \end{tabular}%
    }
  \label{tab:forensicattribute}%
\end{table}%
\textit{\textbf{1) Detection Task:}} Distinguishing real from fake images is the core capability of forensic detectors. We compare \textbf{FakeScope} with other LMMs under the \textit{qualitative} setting using the FakeClass dataset, which includes 6,000 diverse QA pairs with balanced authenticity labels and varied queries. As shown in Table~\ref{tab:fakeclass-result1}, both versions of FakeScope consistently outperform other models across all subcategories, indicating that FakeInstruct notably enhances LMMs’ forensic detection capabilities. FakeScope reaches a peak accuracy of 99.68\%, exceeding even the strongest human expert. Figure~\ref{fakeclass-result2} further compares models under the quantitative setting using calibration-based metrics. FakeScope remains the strongest model, demonstrating the effectiveness of the token-based soft-scoring strategy. Notably, despite receiving no explicit numerical supervision during training, it aligns closely with hard labels and produces reliable probability estimates, suggesting strong internalized reasoning. Even when trained on the same image data, conventional detectors perform worse than FakeScope. Without image-level augmentation, these data-driven detectors degrade markedly~\cite{wang2020cnn}, underscoring the value of FakeInstruct’s multimodal supervision and the stronger forensic integration ability of LMMs.

\textit{\textbf{2) Interpreting Task:}} We evaluate LMMs with the \textit{fault-finding} prompting mode of FakeClue, where models are instructed to explain image authenticity based on visual descriptions, such as in the prompt ``\textit{This is a fake image generated by AI, explain the reasons}''. In the first category of Table~\ref{tab:multimodalcapab}, FakeScope achieve clear gains in this task, which indicates that training on FakeInstruct conspicuously enhances the \textit{effect-to-cause} awareness concerning the forensic attributes of LMMs.

\textit{\textbf{3) Reasoning Task:}} The forensic reasoning task is the inverse of interpretation, assessing LMMs' inference capability in deriving authenticity judgments from visual trace evidence. This task is evaluated using the prompts in the \textit{inference} mode of FakeClue, such as ``\textit{Inspect this image carefully and formulate an evaluation regarding its authenticity}''. As shown in the second category of Table~\ref{tab:multimodalcapab}, baselines generally perform poorly on this task, whereas FakeScope improves substantially after training on FakeInstruct and achieves state-of-the-art results. Most notably, it surpasses the teacher model GPT-4V by more than 30\%, highlighting FakeInstruct’s effectiveness in strengthening cause-to-effect forensic awareness.
 
\textit{\textbf{4) Fine-{G}rained Analyzing Task:}}  We assess FakeScope on FakeQA, which targets fine-grained visual forgery queries. As shown in the third category of Table~\ref{tab:multimodalcapab}, FakeScope clearly outperforms both baseline and teacher models, demonstrating stronger fine-grained forensic analysis. However, this capability still lags behind its interpreting and reasoning performance, suggesting the future potential to improve its answering open-ended questions on subtle forgery artifacts.
\begin{figure*}[!tbp]
\centering
\includegraphics[scale=0.28]{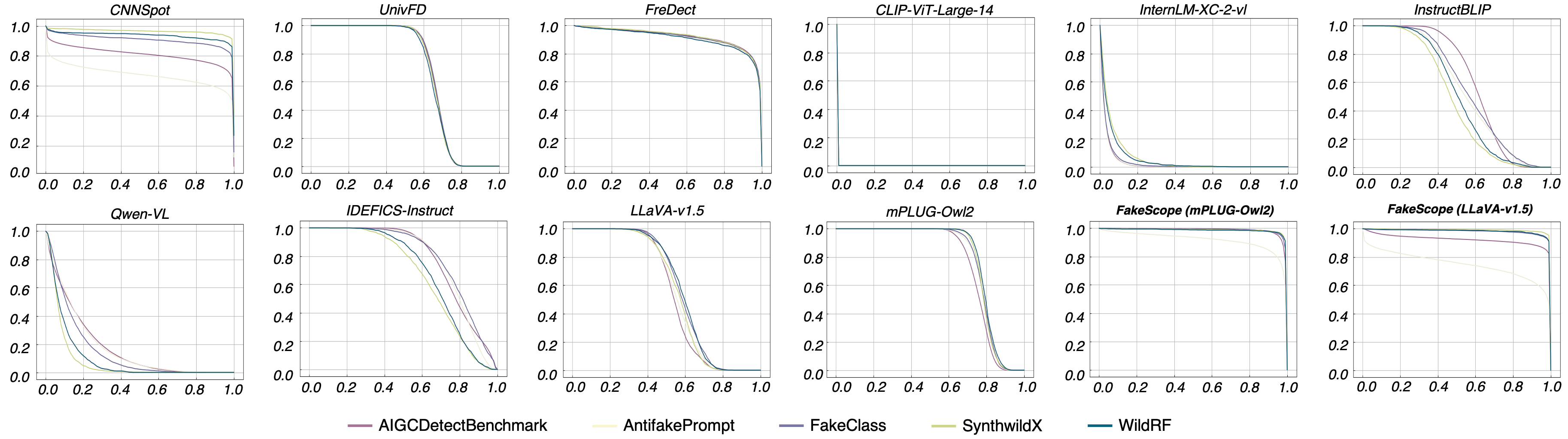}
\caption{Accuracy as a function of detection threshold. For each model, we present the relations between the detection threshold (x-axis) and accuracy (y-axis) on all generators of the five testing databases. The FakeScope of two versions of base models (the last two sub-figures) is the only one that can achieve a great bargain between threshold sensitivity and detection accuracy, indicating remarkable robustness.}
\label{fig:acc-threshold}
\end{figure*}
\begin{table*}[!tbp]
  \centering
  \caption{In-the-wild performance comparison. The AP and AUC (\%) are reported, and top two performers for each category are highlighted in \textbf{bold} and \underline{underline} respectively.}
    \renewcommand{\arraystretch}{1}
  \fontsize{7pt}{8pt}\selectfont %
    \setlength{\tabcolsep}{1.8mm}{
    \begin{tabular}{lccccccc}
    \toprule
\textbf{Datasets} & \multicolumn{3}{c}{\textbf{WildRF}} & \multicolumn{3}{c}{\textbf{SynthWildX}} & \multicolumn{1}{c}{\textbf{Average}} \\
\cmidrule(lr){2-4}\cmidrule(lr){5-7}\cmidrule(lr){8-8}          
\textbf{Models} & \multicolumn{1}{c}{\textit{Reddit}} & \multicolumn{1}{c}{\textit{FB}} & \multicolumn{1}{c}{\textit{X}} &\textit{Dalle3} &\textit{Firefly} &\textit{MJ} & \multicolumn{1}{c}{\textbf{mAP}$\uparrow$/\textbf{AUC}$\uparrow$} \\
    \midrule
     CNNSpot \textit{(training-based)}       &\underline{84.20}/ \underline{82.94}       &\underline{79.58}/ \underline{79.27}       &\underline{86.62}/ \underline{86.46}       &\textbf{91.21}/ \textbf{91.85}       &\underline{83.72}/ \underline{83.39}       &\textbf{87.19}/ \textbf{87.26}       & \underline{85.42}/ \underline{85.20} \\
          UnivFD \textit{(training-based)}       &49.83/ 48.65       & 55.54/ 56.35      &54.89/ 53.67 &41.77/ 38.22 &55.21/ 56.05 &45.88/ 42.83 &50.52/ 49.29          \\
          FreDect \textit{(training-based)} &46.95/ 47.27 &51.97/ 52.62 &48.25/ 48.21 &45.26/ 45.65 &46.77/ 45.38 &50.99/ 52.86 &48.37/ 48.67 \\
          CLIP-ViT-Large-14 \textit{(zero-shot)} &45.64/ 45.52 &38.93/ 34.05 &44.07/ 42.94 &42.33/ 38.19 &42.26/ 38.56 &42.37/ 37.39    &42.60/ 39.44  \\
\arrayrulecolor{gray}    
\cmidrule{1-8}
    InternLM-XC.2-vl \textit{(InternLM2-7B)}     &42.87/ 38.76       &49.93/ 51.08    &42.36/ 38.97  &50.06/ 52.65 &56.39/ 59.33 &48.39/ 49.68 &48.67/ 48.41   \\
    InstructBLIP \textit{(Vicuna-7B)}     &53.75/ 54.82       &40.54/ 34.92       &45.65/ 42.31    &41.24/ 36.38  &42.24/ 38.63  &49.56/ 48.13 &45.83/ 42.53    \\
    Qwen-VL \textit{(Qwen-7B)}     &40.85/ 37.65       &36.46/ 26.29       &35.16/ 22.67       &36.44/ 27.06       &39.15/ 32.64 &34.85/ 20.34 &36.48/ 27.11        \\
    IDEFICS-Instruct \textit{(LLaMA-7B)}     &46.69/ 45.90       &45.96/ 43.65 &41.24/ 34.70    &37.41/ 27.92 &39.93/ 33.36 &40.29/ 32.50 &41.92/ 36.69    \\
    LLaVA-v1.5 \textit{(Vicuna-7B, baseline)}     &58.79/ 62.41       &51.66/ 49.45 &50.67/ 51.70   &53.30/ 56.43 &49.03/ 49.68 &54.70/ 56.45 &53.36/ 54.78     \\
    mPLUG-Owl2 \textit{(LLaMA-7B, baseline)}    &58.52/ 59.66       &51.96/ 50.30       &47.72/ 45.29   &48.23/ 48.05 &46.34/ 48.33 &47.67/ 46.16 &51.74/ 49.63    \\
\cmidrule{1-8}
\arrayrulecolor{black} 
    \textbf{FakeScope} \textit{(\textbf{Ours}, mPLUG-Owl2)} &\textbf{91.89}/ \textbf{91.45}       &\textbf{85.31}/ \textbf{85.43}       & \textbf{92.96}/ \textbf{92.14}      &\underline{89.02}/ \underline{89.43} &\textbf{88.04}/ \textbf{88.56} &\underline{83.72}/ \underline{84.04} &\textbf{88.49}/ \textbf{88.51}       \\
\textbf{FakeScope} \textit{(\textbf{Ours}, LLaVA-v1.5)} &80.01/ 78.60 &67.94/ 73.23              &75.48/ 76.09       & 87.37/ 87.44 &78.94/ 81.76 &81.99/ 83.40 &78.46/ 80.09      \\
    \bottomrule
    \end{tabular}%
    }
  \label{tab:inthewild}%
\end{table*}%
\begin{table*}[!tbp]
  \centering
  \caption{Generalization capability comparison. The AP (\%) of various classifiers across 17 individual generators is reported. Values in {\textcolor{gray}{gray}} represent the generators seen in training. The top performer is highlighted in \textbf{bold}.}
  \renewcommand{\arraystretch}{1}
  \fontsize{7pt}{7pt}\selectfont %
    \setlength{\tabcolsep}{0.4mm}{
    \begin{tabular}{l>{\color{gray}}c>{\color{gray}}cccccc>{\color{gray}}c>{\color{gray}}c>{\color{gray}}c>{\color{gray}}ccccccccc}
    \toprule
    \textbf{Datasets} &\multicolumn{6}{c}{\textbf{AIGCDetectBenchmark}}&\multicolumn{11}{c}{\textbf{AntifakePrompt}}  &\multirow{2}*{\textbf{Unseen}} &\textbf{Average}\\
    \cmidrule(lr){1-1} \cmidrule(lr){2-7} \cmidrule(lr){8-18} \cmidrule(lr){20-20}
    \textbf{Models} &\textit{\textcolor{gray}{Pro}} &\textit{\textcolor{gray}{Style}} &\textit{Star} &\textit{Gau} &\textit{Style2} &\textit{WFIR} &\textit{Control}  &\textit{\textcolor{gray}{SD2}} & \textit{\textcolor{gray}{SDXL}}  & \textit{\textcolor{gray}{IF}} & \textit{\textcolor{gray}{Dalle2}} &\textit{SGXL} &\textit{LaMa} &\textit{SD2Inpaint} &\textit{LTE} &\textit{SD2SR} &\textit{DeepFake}&&\textbf{mAP}$\uparrow$\\
    \midrule
    CNNSpot~\cite{wang2020cnn} & 64.86      & 61.41      & 70.33             & 60.10      & 63.08      & 56.41      &70.19 &82.85 &92.51 &86.50 &56.61 &60.25 &48.88 &52.15 &50.54 &50.22 &95.15 &60.71 &66.00  \\
    UnivFD~\cite{ojha2023towards} &61.37       &50.36       &39.78              &55.61       &48.49       &51.91       &56.22 &43.25 &35.75 &44.24 &45.36 &58.17 &50.85 &57.46 &80.99 &60.68 &68.42 &57.24 &53.47 \\
    FreDect~\cite{frank2020leveraging} &49.93  &55.13  &49.99 &53.26 &54.93 &49.97 &53.61 &29.79 &37.78  &38.36 &35.80 &44.46 &49.86 &49.93 &48.02 &49.81 &56.10 &50.63 &47.45 \\
    CLIP-ViT-Large-14~\cite{radford2021learning} &89.22              &71.12       &94.92       &88.58       &64.40       &81.17       &60.43 &29.20 &31.87 &37.28 &49.44 &70.85 &62.86 &63.26 &85.15 &71.71 &41.85 &71.38 &64.31  \\
\arrayrulecolor{gray}    
\midrule
    InternLM-XC.2-vl~\cite{dong2024internlm}     & 52.15             &46.40       & 61.16      &50.47       & 52.28      &44.29       &68.72 &35.73 &41.03 &48.59 &45.49 &62.15 &52.39 &58.33 &60.58 &58.46 &71.70 &58.23 &53.52  \\
    InstructBLIP~\cite{li2023blip}     &50.27              &40.79       &32.31       &88.63       &40.42       &49.36       &55.46 &25.98 &29.87 &25.86 &72.11 &64.04 &59.18 &74.46 &95.64 &81.84 & 97.48 &67.17 &57.86  \\
    Qwen-VL~\cite{bai2023qwen}            &72.27       &42.91       &41.39       &82.54       &44.80       &48.92       &84.18 &75.39 &69.55 &72.97 &83.83 &88.15 &66.50 &78.31 &93.55 &89.05 &91.31 &73.52 &72.10  \\
    IDEFICS-Instruct~\cite{laurenccon2023obelics}     &40.26       &51.87       &53.68             &39.40       &47.26       &66.99       &35.16 &22.03 &23.55 &25.85 &22.45 &41.24 &46.53 &41.11 &34.81 &44.27 &48.49 &45.36 &40.29  \\
    mPLUG-Owl2~\cite{ye2023mplug}    &58.44       &48.91       &83.24       &51.39       &45.26          &41.84       &69.79 &45.10 &46.10 &44.71 &55.03 &74.34 &50.71 &57.82 &77.64 &68.28 &71.16 &62.86 &58.22 \\
    LLaVA-v1.5~\cite{liu2024visual}    &40.97       &44.09       &41.30       &38.08       &43.03       &56.01            &42.36 &24.32 &25.07 &26.71 &32.78 &45.13 &47.93 &45.08 &37.69 &43.91 &80.87 &47.40 &42.08  \\
\midrule
    \textbf{FakeScope} \textit{(mPLUG-Owl2)}&89.70       &92.96       &88.63       &89.83       &86.94           &93.32       &98.20 &97.27 &\textbf{99.17} &98.80 &\textbf{99.04} &\textbf{99.77} &\textbf{98.20} &\textbf{98.99} &\textbf{99.72} &\textbf{98.80} &\textbf{99.96} &\textbf{95.67} &95.84 \\
    \textbf{FakeScope} \textit{(LLaVA-v1.5)}  &\textbf{99.92}       &\textbf{98.01}       &\textbf{100.00}       &\textbf{99.66}       &\textbf{98.01}       &\textbf{99.45}       &\textbf{98.96} &\textbf{98.21} &99.04 &\textbf{99.10} &\textbf{99.04} &99.65 &82.59 &95.97 &96.40 &97.20 &73.64 &94.68 &\textbf{96.17}  \\
    \arrayrulecolor{black}
    \bottomrule
    \end{tabular}%
    }
  \label{tab:generalization}%
\end{table*}%

\textbf{\textit{5) Forensic Attribution Task:}} We evaluate forensic attribution on FakeBench\textsuperscript{\texttt{MCQ}}, containing close-ended questions on specific visual evidence. As listed in Table~\ref{tab:forensicattribute}, FakeScope achieves clear improvements over its base models, indicating that it can not only justify detection with long-form rationales but also identify concrete forensic evidence effectively. This further confirms that FakeInstruct successfully imparts visual forensic awareness to LMMs. Among all dimensions, the largest gain appears on \textit{where} questions. We also observe larger gains on low- and medium-level attributes, suggesting the model could potentially be further improved by incorporating more conceptually high-level forensic knowledge.
\subsection{In-the-wild Performance}
\label{exp5}
To assess real-world applicability, we evaluate on in-the-wild datasets built from social-media imagery, including WildRF~\cite{cavia2024real} and SynthWildX~\cite{cozzolino2024raising}. These contain real-world degradations which pose additional challenges for forensic models. As reflected by the results in Table~\ref{tab:inthewild}, FakeScope equipped with mPLUG-Owl2~\cite{ye2023mplug} consistently outperforms prior state-of-the-art methods across most generators and achieves the best overall in-the-wild performance. It is also competitive on Dalle3 and Midjourney, ranking second only to CNNSpot~\cite{wang2020cnn}. 
These findings highlight FakeScope’s robustness in handling real-world forensic challenges beyond in-lab benchmarks.
\subsection{Generalization Capability}
\label{exp3}
The generalization to unacquainted image contents and generators is crucial for practical deployment. To assess this, we test all methods on the uncontaminated content proportions of AntifakePrompt~\cite{chang2023antifakeprompt} and AIGCDetectBenchmark~\cite{zhong2024patchcraft}, whose content is disjoint from the data used in FakeInstruct. AntifakePrompt further includes OOD tasks beyond our primary scope, such as image super-resolution, face spoofing~\cite{roessler2019faceforensicspp}, and inpainting, which stress both robustness and generalization. Table~\ref{tab:generalization} presents results across 17 image generators, including six unseen and 11 unseen ones. Both FakeScope variants achieve superiority across all 17 generators and outperform all baselines on the unseen subset, indicating strong generalization to novel content and generation techniques.

We further evaluate stability by plotting detection accuracy as a function of the decision threshold ($\in (0,1)$), measuring sensitivity to threshold selection across generators. In Fig.~\ref{fig:acc-threshold}, each curve is obtained by averaging accuracy over all generators in each dataset. Unlike AUC and mAP, accuracy is highly threshold-dependent; unstable detectors can degrade substantially under suboptimal thresholds. As shown in Fig.~\ref{fig:acc-threshold} both FakeScope variants exhibit consistent peak accuracy and remain stable over a wide threshold range. In contrast, competing methods show larger shifts in their optimal thresholds and greater performance volatility. This suggests FakeScope is less sensitive to threshold choice across datasets and generators, improving robustness in real-world settings.

\subsection{Ablation Studies}
\label{exp4}
We analyze how data composition and scale affect FakeScope training with mPLUG-Owl2~\cite{ye2023mplug}. Specifically, we study: (1) instruction scale, i.e., the effect of dataset size; (2) instructional diversity, by comparing FakeChain-only training against enriched multimodal supervision from FakeInstruct; (3) multimodal awareness, via detection performance with and without multimodal context. Besides, we further validate the (4) effectiveness of the proposed token soft-scoring strategy.

\begin{figure}[!tbp]
\centering
\includegraphics[scale=0.9]{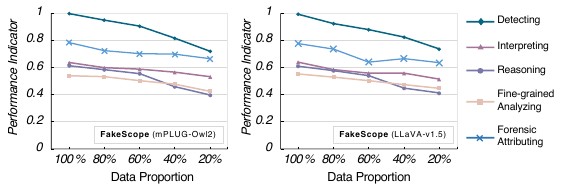}
\caption{Performance on FakeBench~\cite{li2024fakebench} concerning training data scale. The x-axis indicates the proportion of FakeInstruct used for training, and y-axis shows corresponding performance indicator of each ability dimension.}
\label{fig:fewshot}
\end{figure}

\textit{\textbf{1) Effects of Training Data Scale}}: Data scarcity is a core bottleneck for forensic modeling because high-quality annotations are expensive to obtain. To assess the data efficiency of \textbf{FakeInstruct}, we sample 80\%, 60\%, 40\%, and 20\% subsets while preserving the original composition. As shown in Fig.~\ref{fig:fewshot}, FakeScope degrades gracefully as training data decreases, indicating strong robustness of FakeInstruct. Even at 20\%, performance remains clearly above baselines, and improves steadily with larger subsets. This trend suggests that the gains are driven primarily by the forensic knowledge encoded in FakeInstruct rather than data volume alone. 


\textit{\textbf{2) Effects of Instructional Diversity}}: We jointly train a unified forensic foundation model on multiple subsets and compare this setting with task-specific training focused only on transparent forensics. As shown in Table~\ref{tab:ablation1}, collaborative training that combines long-chain reasoning with fine-grained visual instruction consistently improves performance, which benefits cross-task generalization. Full-schedule training improves all capabilities while reducing cross-task interference, indicating complementary rather than competitive interactions among forensic knowledge types. In particular, both analysis and attribution benefit from fine-grained supervision, whereas training with FakeChain alone provides no comparable improvement. These results validate the effectiveness of our visual instruction strategy.

\begin{table}[!tbp]
  \centering
  \caption{Comparison on transparent forensic capabilities between \textit{only} \textbf{FakeChain} and \textit{full} \textbf{FakeInstruct} as training dataset. The best performance of each column is highlighted in \textbf{bold}.}
  \renewcommand{\arraystretch}{1}
  \fontsize{7pt}{7pt}\selectfont %
    \setlength{\tabcolsep}{1.5mm}{
    \begin{tabular}{lccccc}
    \toprule
    Training Strategy &Detect & Interpret & Reason & Analyze & Attribute\\
    \midrule
    none (\textit{baseline}) &58.22\% &0.423 &0.296 &0.368 &65.97\%\\
    \arrayrulecolor{gray}
    \midrule
    \arrayrulecolor{black}
    \textit{only} \textbf{FakeChain}  &83.27\%       &0.593    &0.576   &0.350  &68.43\%\\
    \textit{full} \textbf{FakeInstruct}  &\textbf{99.68\%}       & \textbf{0.638}      &\textbf{0.614}  &\textbf{0.540} &\textbf{78.50\%}\\
    \bottomrule
    \end{tabular}%
    }
  \label{tab:ablation1}%
\end{table}%
\begin{table}[!tbp]
  \centering
  \caption{Comparison on forensic detection ability among \textit{full} \textbf{FakeInstruct}, \textit{only} absolute judgment, and \textit{only} contextual data containing trace evidence instructions. AP (\%) on each dataset is reported.}
  \renewcommand{\arraystretch}{1}
  \fontsize{7pt}{8pt}\selectfont %
    \setlength{\tabcolsep}{1mm}{
    \begin{tabular}{lcccc}
    \toprule
    Dataset/ Training Strategy &\makecell{none\\(\textit{baseline})} &\makecell{\textit{only}\\absolute} &\makecell{\textit{only}\\contextual} &\makecell{\textit{full}\\(\textit{employed})} \\
    \midrule
    FakeBench~\cite{li2024fakebench} &68.37 &94.06$_\text{\textcolor{mygreen}{+25.96}}$ &84.13$_\text{\textcolor{mygreen}{+15.76}}$       &\textbf{99.94}$_\text{\textcolor{mygreen}{+31.57}}$  \\
    AntiFakePrompt~\cite{chang2023antifakeprompt} &60.06 &90.12$_\text{\textcolor{mygreen}{+30.06}}$       &74.84$_\text{\textcolor{mygreen}{+14.78}}$       &\textbf{98.90}$_\text{\textcolor{mygreen}{+38.84}}$  \\
    AIGCDetecBenchmark~\cite{zhong2024patchcraft} &54.84 &81.65$_\text{\textcolor{mygreen}{+26.81}}$ &60.73$_\text{\textcolor{mygreen}{+5.89}}$       &\textbf{90.23}$_\text{\textcolor{mygreen}{+35.39}}$  \\
    WildRF~\cite{cavia2024real} &52.73 &82.34$_\text{\textcolor{mygreen}{+29.61}}$ &62.54$_\text{\textcolor{mygreen}{+9.81}}$       &\textbf{90.05}$_\text{\textcolor{mygreen}{+37.32}}$  \\
    SynthWildX~\cite{cozzolino2024raising} &47.41 &82.78$_\text{\textcolor{mygreen}{+35.37}}$  &52.59$_\text{\textcolor{mygreen}{+5.18}}$       &\textbf{86.93}$_\text{\textcolor{mygreen}{+39.52}}$  \\
    \bottomrule
    \end{tabular}%
    }
  \label{tab:ablation2}%
\end{table}%

\textit{\textbf{3) Effects of Multimodal Awareness}}: A key distinction between \textbf{FakeScope} and conventional binary detectors is the use of multimodal evidence for visual trace analysis. Although this design is intended to improve transparent forensics, its contribution to basic detection is not obvious. We therefore compare models trained with and without trace-evidence instructions. As reported in Table~\ref{tab:ablation2}, adding multimodal context improves both detection accuracy and generalization, even under the same image budget. By contrast, removing absolute authenticity labels yields less stable gains. These results indicate that multimodal forensic cues enhance task awareness and fine-grained visual understanding, whereas robust detection still requires explicit authenticity supervision.
\begin{table}[!tbp]
\renewcommand{\arraystretch}{1.1}
\fontsize{6pt}{7pt}\selectfont %
  \centering
  \caption{Ablations of the proposed token soft-scoring strategy: (i) auxiliary classifier; (ii) verbalized confidence; (iii) synonym ensembles for \textless \textit{auth\_token}\textgreater; and (iv) \textit{argmax} vs. \textit{softmax} scoring. We report mAP (\%) on each dataset.}
  \setlength{\tabcolsep}{0.4mm}{
    \begin{tabular}{cllccccc}
    \toprule
    \textit{Exp.} & Model & Strategy &\makecell[c]{Fake\\Bench} & \makecell[c]{Antifake\\Prompt} & \makecell[c]{AD\\Bench.} & WildRF & \makecell[c]{Synth\\WildX} \\
    \hhline{--------}
    \multirow{4}[4]{*}{\textit{(i)}} & \multirowcell{2}[0pt][l]{FakeScope\\(mPLUG-Owl2)} & \gray{\textit{coupled} (\textit{ours})} & \gray{\textbf{99.94}}      & \gray{\textbf{98.90}}      & \gray{\underline{90.23}}      & \gray{\textbf{90.05}}      &\gray{\textbf{86.93}}     \\
          &      & \textit{de-coupled} & 98.67      & 95.20      & 87.69      & \underline{85.42}      & \underline{83.89}        \\
\hhline{~-------}          & \multirowcell{2}[0pt][l]{FakeScope\\(LLaVA-v1.5)} & \gray{\textit{coupled} (\textit{ours})} &\gray{\underline{99.39}}       & \gray{\underline{94.53}}     & \gray{\textbf{99.15}}      & \gray{74.48}      & \gray{82.77}        \\
          &       & \textit{de-coupled} & 96.78      & 95.99      & 96.45      & 60.80      & 79.76        \\
\hhline{--------}    \multirow{8}[2]{*}{\textit{(ii)}} & \multirowcell{4}[0pt][l]{FakeScope\\(mPLUG-Owl2)} & \gray{\textit{token-based} (\textit{ours})}  & \gray{\textbf{99.94}}      & \gray{\textbf{98.90}}      & \gray{\underline{{90.23}}}      & \gray{\textbf{90.05}}      &\gray{\textbf{86.93}}       \\
          &       & \textit{verbalize\textsubscript{greedy}} & 66.08      & 60.43      & 50.87      & 52.36      & 45.42       \\
          &       & \textit{verbalize\textsubscript{t=0.5}} & 66.79      & 61.66      & 52.80      & 52.39      & 45.40      \\
          &       & \textit{verbalize\textsubscript{t=1.0}} & 65.32      & 58.75      & 51.43      & 50.28      & 43.27    \\
          \hhline{~-------}& \multirowcell{4}[0pt][l]{FakeScope\\(LLaVA-v1.5)} & \gray{\textit{token-based} (\textit{ours})}  &\gray{\underline{99.39}}       & \gray{\underline{94.53}}     & \gray{\textbf{99.15}}      & \gray{\underline{74.48}}      & \gray{\underline{82.77}}     \\
          &       & \textit{verbalize\textsubscript{greedy}} & 68.89      & 62.58      &52.66       & 54.90      & 49.87\\
          &       & \textit{verbalize\textsubscript{t=0.5}} & 68.72      & 62.05      & 51.43      & 54.95      & 49.26       \\
          &       & \textit{verbalize\textsubscript{t=1.0}} & 66.77      & 61.02      & 51.35      & 52.73      & 46.58       \\
    \hhline{--------}
    \multirow{12}[3]{*}{\textit{(iii)}} & \multirowcell{6}[0pt][l]{FakeScope\\(mPLUG-Owl2)} & \gray{\textit{fake}$\leftrightarrow$\textit{real} (\textit{ours})} & \gray{\textbf{99.94}}      & \gray{\textbf{98.90}}      & \gray{90.23}      & \gray{\textbf{90.05}}      &\gray{\textbf{86.93}}        \\
          &       & \textit{artificial}$\leftrightarrow$\textit{genuine} & 89.68      & 88.43      & 81.97      & 80.55      & 75.49       \\
          &       & \textit{synthetic}$\leftrightarrow$\textit{authentic} & 92.90      & 92.36      &  84.43     & 87.96      & 82.75       \\
          &       & \textit{fake+artificial}$\leftrightarrow$\textit{real+genuine} & 93.38      & 92.74      & 84.69      & 86.08      & 86.73       \\
          &       & \textit{fake+synthetic}$\leftrightarrow$\textit{real+authentic} & 93.29      &  91.40     & 84.92      & 85.48      &  83.12       \\
          &       &\makecell[l]{%
\textit{fake+artificial+synthetic}\\
$\leftrightarrow$
\textit{real+genuine+authentic}%
} & 97.78      & \underline{97.29}      & 90.30      & \underline{89.03}      & \underline{86.81}        \\
\hhline{~-------}          & \multirowcell{6}[0pt][l]{FakeScope\\(LLaVA-v1.5)} & \gray{\textit{fake}$\leftrightarrow$\textit{real} (\textit{ours})} &\gray{\underline{99.39}}       & \gray{94.53}     & \gray{\textbf{99.15}}      & \gray{74.48}      & \gray{82.77}     \\
          &       & \textit{artificial}$\leftrightarrow$\textit{genuine} &  89.53     & 88.65      & 83.77      & 70.64      & 78.65        \\
          &       & \textit{synthetic}$\leftrightarrow$\textit{authentic} & 90.92      & 87.31      & 90.65      & 68.19      & 76.44       \\
          &       & \textit{fake+artificial}$\leftrightarrow$\textit{real+genuine} & 94.92      & 93.31      & 97.65      & 71.19      & 81.49       \\
          &       & \textit{fake+synthetic}$\leftrightarrow$\textit{real+authentic} & 99.42      & 92.57      & 98.23      & 70.62      & 81.68       \\
          &       & \makecell[l]{%
\textit{fake+artificial+synthetic}\\
$\leftrightarrow$
\textit{real+genuine+authentic}%
} & 99.43      & 92.88      & \underline{99.02}      &72.65       &  83.21      \\
    \hhline{--------} \multirow{4}[2]{*}{\textit{(iv)}} & \multirowcell{2}[0pt][l]{FakeScope\\(mPLUG-Owl2)} & \gray{\textit{softmax} (\textit{ours})} & \gray{\textbf{99.94}}      & \gray{\textbf{98.90}}      & \gray{\underline{90.23}}      & \gray{\textbf{90.05}}      &\gray{\textbf{86.93}}      \\
          &       & \textit{argmax} & 95.66      &  \underline{98.47}     & 63.57      & 55.36 & 78.78             \\
\hhline{~-------}          & \multirowcell{2}[0pt][l]{FakeScope\\(LLaVA-v1.5)} & \gray{\textit{softmax} (\textit{ours})} &\gray{\underline{99.39}}       & \gray{94.53}     & \gray{\textbf{99.15}}      & \gray{\underline{74.48}}      & \gray{\underline{82.77}}      \\
          &       & \textit{argmax} & 93.55      & 95.64      & 85.55      & 52.92      & 77.42     \\
    \bottomrule
    \end{tabular}%
    }
  \label{tab:ablation4}%
\end{table}%

\textit{\textbf{4) Effects of token soft-scoring}}: We assess the token soft-scoring strategy via four ablations. \textit{(i)} We train an auxiliary classifier by attaching a linear head to the language output and optimizing it on training images, and compare it with our training-free scoring. \textit{(ii)} We evaluate verbalized confidence by prompting the model to output a probability: ``\texttt{<Image> Give a probability between zero and one for the image being fake}''. \textit{(iii)} We test different synonym ensembles for \textless \textit{auth\_token}\textgreater\ to measure lexical sensitivity. \textit{(iv)} We compare hard \textit{argmax} (\textit{fake}$\leftrightarrow$\textit{real}) against the proposed \textit{softmax} formulation. The results are listed in part \textit{(i)} to \textit{(iv)} of Table~\ref{tab:ablation4}. Token soft-scoring achieves the best overall accuracy and stability across settings. The auxiliary classifier is competitive but introduces extra parameters and training. Anchor-token substitution causes only moderate variation, much smaller than backbone-dependent performance gaps. Verbalized confidence performs worst on all datasets and is highly sensitive to decoding temperature, leading to unstable outputs, whereas token soft-scoring is insensitive to decoding settings. Although \textit{argmax} is comparable on a few datasets, its generalization is weaker.
\section{Conclusions and Future Work}
This paper proposes a multimodal paradigm for transparent AI-generated image forensics that integrates detection, in-depth analysis, and extended discussions within a unified framework. To support this paradigm, we build FakeChain, a large-scale forensic reasoning dataset created through human-machine collaboration, and extend it into FakeInstruct, the first multimodal instruction-tuning dataset for image forensics with 2 million visual instructions. On this basis, we develop FakeScope, an expert large multimodal model that achieves superior performance in both closed-ended and open-ended forensic tasks, while showing strong generalization and robustness in the wild. Notably, FakeScope demonstrates zero-shot quantitative detection without explicit numerical supervision, suggesting an emergent capacity for probabilistic reasoning.

More broadly, this work offers a practical path toward transparent and trustworthy AI-generated image forensics. Beyond accurate detection, FakeScope produces cross-modal forensic information, supporting explainable analysis. Our results also underscore the value of human-in-the-loop design and structured prompting for reliable knowledge distillation in LMMs. Although current progress still depends on advanced models such as GPT-4V and important uncertainties remain, the findings confirm the value of multimodal information for forensic transparency. Looking ahead, we intend to explore the abilities of newly emerged LMMs and contribute model updates to the community. Given the diminishing returns implied by scaling laws, future efforts may focus on enhancing few-shot adaptability and dialectical reasoning to boost robustness in more complex, unseen forensic scenarios. 
{
\bibliographystyle{IEEEtran}
\bibliography{refs}
}

\end{document}